\documentclass[11pt]{article}

\usepackage[final]{acl}

\usepackage{times}
\usepackage{latexsym}

\usepackage[T1]{fontenc}

\usepackage[utf8]{inputenc}

\usepackage{microtype}

\usepackage{inconsolata}

\usepackage{graphicx}
\usepackage{booktabs}
\usepackage{multicol}
\usepackage{lipsum}
\usepackage{amsmath}
\usepackage{tabularx}
\usepackage[table]{xcolor}
\definecolor{poscolor}{RGB}{222,235,247}  
\definecolor{lencolor}{RGB}{230,245,208}  
\definecolor{patcolor}{RGB}{252,228,236}  
\definecolor{extcolor}{RGB}{229,229,229}  

\title{VisDoT : Enhancing Visual Reasoning through Human-Like Interpretation Grounding and Decomposition of Thought }

\author{
  Eunsoo Lee$^{1}$,
  Jeongwoo Lee$^{2}$,
  Minki Hong$^{1}$,
  Jangho Choi$^{1}$,
  Jihie Kim$^{1\dagger}$
  \\
  $^{1}$Department of Computer Science and Artificial Intelligence, Dongguk University \\
  $^{2}$Department of Electronics and Electrical Engineering, Dongguk University \\
  \texttt{\{dmstn7432, jwlee0519, jackyh1, 2025120382\}@dgu.ac.kr, jihie.kim@dgu.edu}
  \\
}

\begin{document}
\maketitle
\begin{abstract}
Large vision-language models (LVLMs) struggle to reliably detect visual primitives in charts and align them with semantic representations, which severely limits their performance on complex visual reasoning. This lack of perceptual grounding constitutes a major bottleneck for chart-based reasoning. We propose VisDoT, a framework that enhances visual reasoning through human-like interpretation grounding. We formalize four perceptual tasks based on the theory of graphical perception such as position and length. Building on this foundation, we introduce decomposition-of-thought (DoT) prompting, which sequentially separates questions into visual perception sub-questions and logic sub-questions. Fine-tuning InternVL with VisDoT achieves a +11.2\% improvement on ChartQA and surpasses GPT‑4o on the more challenging ChartQAPro benchmark. On the newly introduced VisDoTQA benchmark, the model improves by +33.2\%. Furthermore, consistent zero-shot gains on diverse open-domain VQA benchmarks confirm the generalizability of the perception-logic separation strategy for visual question answering in general. VisDoT leverages human-like perception to enhance visual grounding, achieving state-of-the-art chart understanding and interpretable visual reasoning.
\end{abstract}

\begin{figure}[!t]
    \centerline{\includegraphics[width=\columnwidth]{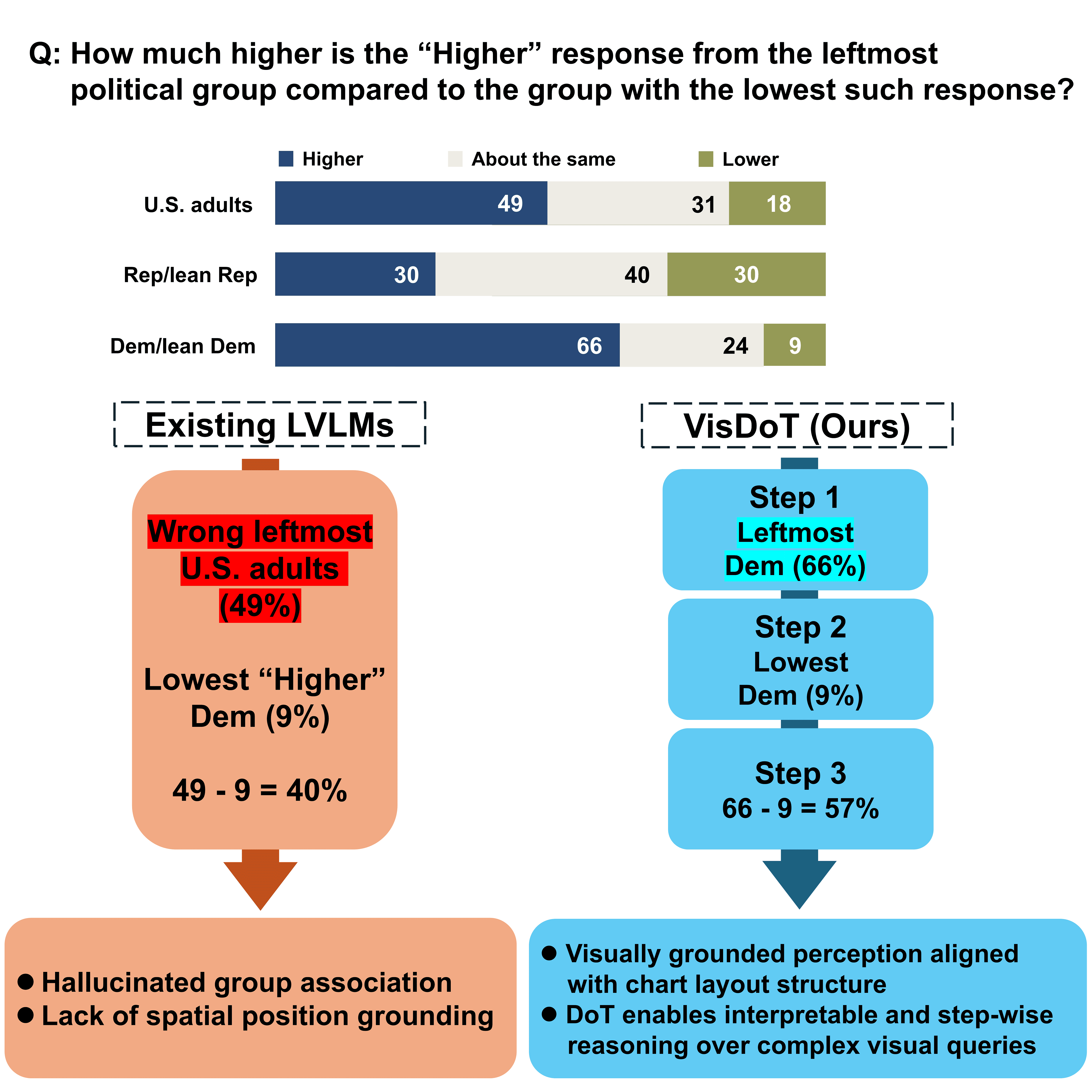}}
    \caption{Comparison of visual reasoning between existing LVLMs and VisDoT. While LVLMs fail to visual perception in spatial structure, VisDoT leverages decomposition-of-thought (DoT) to accurately infer answers through sequential visual analysis.}
    \label{fig:example} 
\end{figure}


\section{Introduction}
\label{Sec:1}
One of the fundamental challenges for vision‑language models (VLMs) is their limited ability to interpret Visualized Data such as charts, graphs, infographics, and dashboards and to perform reasoning over interrelated visual elements. In real‑world scenarios, large vision‑language models (LVLMs) often fail to reliably detect visual primitives and align them with semantic representations when users issue queries or instructions about visual data without explicitly mentioning identifiers such as legend labels or axis names. This results in substantial performance degradation (Figure~\ref{fig:example}). Prior work on Visualized Data understanding has focused primarily on adapting VLMs through instruction tuning and chain‑of‑thought (CoT) supervision\cite{meng2024chartassisstantuniversalchartmultimodal, han2023chartllamamultimodalllmchart}. These approaches are typically limited to simple text-based keyword–value mappings and lack robust grounding mechanisms that align visual primitives such as color, spatial coordinates, and shape with their corresponding real-world entities. As a result, their performance degrades substantially on tasks that require high-level perceptual alignment, such as legend identification and multi-object comparison \cite{masry2022chartqa, masry-etal-2025-chartqapro}. Similarly, while CoT supervision has shown effectiveness for text-only reasoning, it provides only limited gains in visual reasoning contexts that require logical reasoning \cite{cuarbune2024chart, wu2025grounded}.

In contrast, recent studies have demonstrated the potential of LVLMs for visual question answering (VQA) by integrating visual elements into instruction-following datasets~\cite{masry-etal-2025-chartgemma}. In parallel, research on modular VLMs has shown that disentangling reasoning, visual understanding, and language understanding is an effective strategy for VQA~\cite{amizadeh2020neuro}. Furthermore, recent work suggests that a similar separation between visual perception and logical reasoning is also necessary for LVLMs~\cite{zhang2024far}.

Based on these insights, we pose two central research questions: (1) \textit{How can LVLMs be effectively adapted for visualized data-based question answering?}, and (2) \textit{Can task decomposition strategies be effectively extended to LVLMs?}

To address these, we propose VisDoT, a framework that enhances the grounding and reasoning capabilities of LVLMs. VisDoT leverages graphical perception theory to improve visual grounding and employs systematic task decomposition to enable structured reasoning. 
Building on the theory of human graphical perception \cite{cleveland1984graphical}, we formalize core perceptual tasks: Position, Length, Pattern, and Extract. These tasks serve as a foundation for aligning model attention with human perceptual principles.
We further introduce a decomposition-based prompting strategy, decomposition-of-thought (DoT), which separates questions into perceptual and logical reasoning stages to enable sequential problem solving. The LVLM is trained to perform the four perceptual tasks and to adopt the DoT reasoning strategy.

A model trained with the proposed framework provides two key benefits:
(1) improved robustness to complex and underspecified visual queries, and (2) enhanced interpretability of VLM failure cases. Experiments on ChartQA~\cite{masry2022chartqa}, ChartQAPro~\cite{masry-etal-2025-chartqapro}, and cross-domain visual reasoning benchmarks~\cite{li2023evaluating, yue2024mmmu} yield state-of-the-art performance, underscoring the framework’s effectiveness for real-world visual reasoning. In particular, the model achieves an 11.2\% gain on ChartQA and matches or surpasses GPT-4o~\cite{openai2024gpt4o} on the more challenging ChartQAPro and our VisDoTQA benchmark. Notably, applied to the open‑domain VQA benchmarks POPE~\cite{li2023evaluating} and MMMU~\cite{yue2024mmmu}, our DoT prompting raises performance by 1.43\% and 2.2\% absolute over an identical CoT‑trained backbone under the same compute budget. By requiring each sub‑question to first localise the relevant visual element before reasoning, DoT removes the grounding errors that limit CoT, showing that the approach generalises well beyond chart tasks.

Our contributions are summarized as follows:
\begin{itemize}
    \item We formalize four core perceptual tasks grounded in graphical perception theory, establishing the foundation for modeling LVLMs that emulate human decoding of data visualizations.
    \item We introduce a novel DoT strategy that enables LVLMs to emulate human-like visual interpretation by separating complex questions into perceptual and logical steps.
    \item We construct a perception-following dataset that combines these perceptual tasks with DoT-based prompting, enabling chart understanding, visual–linguistic grounding, and compositional reasoning.
    \item Our trained models achieve state-of-the-art performance across multiple visual reasoning benchmarks including VQA tasks, demonstrating the effectiveness of the VisDoT framework in grounding visual-language inference.
\end{itemize}

\section{Related Work}
\label{Sec:2}


\begin{figure*}[!t]
    \centerline{\includegraphics[width=\textwidth]{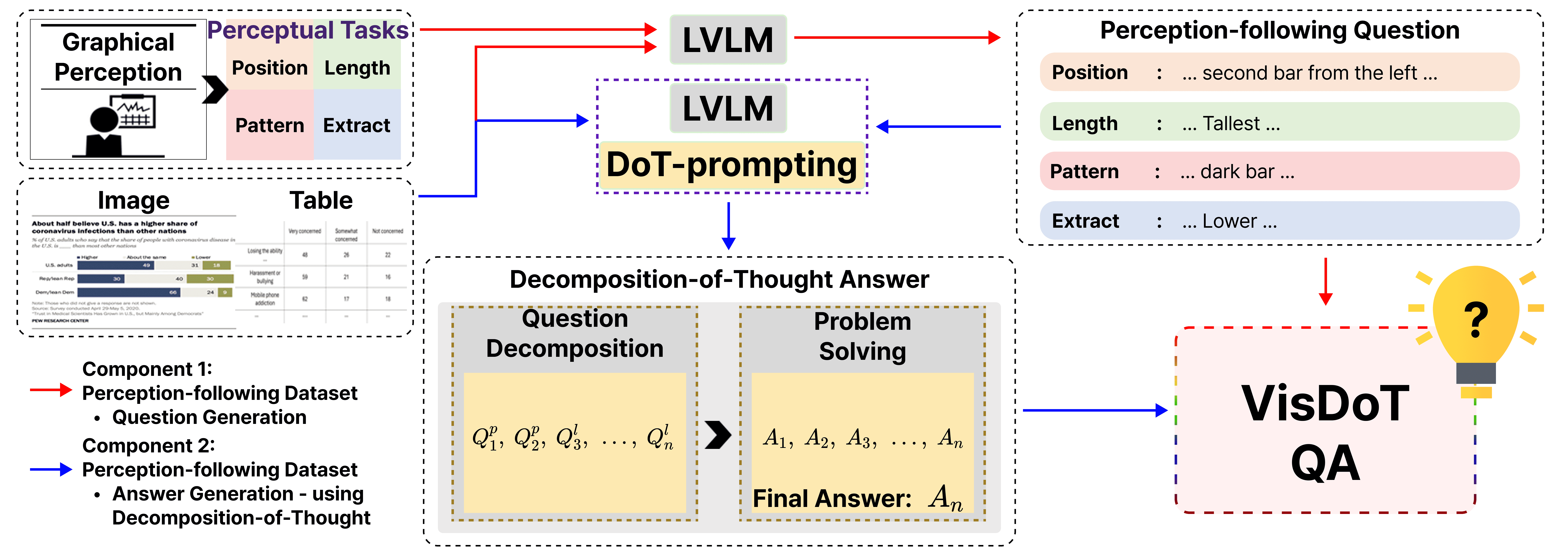}}
    \caption{An overview of our framework.}
    \label{fig:framework_overview}
\end{figure*}


\subsection{Reasoning from Visualized Data}
\label{Sec:2.1}
Recent advances in LVLMs have spurred a growing body of research aimed at improving structured visual reasoning for complex inputs such as charts. Pix2Struct~\cite{lee2023pix2struct} improves layout understanding through markup-based reconstruction pretraining, while Matcha~\cite{liu-etal-2024-enhancing} focuses on chart de-rendering and numerical reasoning.
Alongside the widespread adoption of instruction tuning, several studies have proposed adapting LVLMs to visualized data understanding by leveraging synthetic or large-scale instruction datasets~\cite{han2023chartllamamultimodalllmchart, masry2024chartinstruct, masry-etal-2025-chartgemma}. More recently, ChartGemma~\cite{masry-etal-2025-chartgemma} attempts to enhance the visual capabilities of LVLMs by incorporating a visual-element-based instruction tuning dataset into the training data.
Chart-based Reasoning~\cite{cuarbune2024chart} reports improved VQA performance by generating intermediate reasoning processes or structured evidence. Similarly, prior work~\cite{han2023chartllamamultimodalllmchart, masry2024chartinstruct, masry-etal-2025-chartgemma} improves chart reasoning performance by augmenting training datasets with CoT annotations. However, recent findings~\cite{zhang2024far} emphasize that CoT~\cite{wei2022chain} and Self-Consistency~\cite{wang2022self} have limited effectiveness in vision tasks.

\subsection{Decomposition Strategies for Complex Visual Reasoning}
\label{Sec:2.2}
To address complex visual reasoning, prior work has proposed various decomposition strategies, including modular execution and question decomposition. As a representative early approach, NS-VQA~\cite{amizadeh2020neuro} resolves this challenge by modularizing recognition and inference through a neuro-symbolic architecture.
More recently,~\cite{wang2023filling, hu2023avis} improve VQA performance by invoking executable programs and tools or leveraging multiple agents. However, iterative modular execution incurs high computational costs and makes it difficult to transfer the approach to a single end-to-end sVLM.
Question decomposition methods~\cite{khan2023exploring, zhang2024visual} suggest that splitting complex queries into sub-questions can improve LVLM performance. However,~\cite{khan2023exploring} does not handle a large number of sub-questions effectively, and~\cite{zhang2024visual} mainly focuses on demonstrating the potential of question decomposition strategies in LVLMs. In contrast, VisDoT integrates question decomposition and reasoning within a single system and a single reasoning process. This enables scalable annotation generation and effective knowledge transfer to compact LVLMs.


\begin{table}[t]
\small
\centering
\begin{tabularx}{\linewidth}{@{}l X@{}}
\toprule
\textbf{Task} & \textbf{Description} \\
\midrule
Position & Compares object positions along a common scale (e.g., x- or y-axis) to determine relative order. As the most accurate perceptual channel for conveying quantitative information, it plays a critical role in effective visual communication. \\
Length   & A distortion-free visual attribute across visualized data types~\cite{stevens2017psychophysics}, used as a secondary cue to position. \\
Pattern  & Links pattern cues to legends and data to distinguish categories. Assesses visual label mapping ability. \\
Extract  & Reads explicitly shown values. Evaluates numerical recognition similar to standard QA tasks. \\
\bottomrule
\end{tabularx}
\caption{Core perceptual tasks derived from graphical perception theory, designed to align LVLM attention with structured visual features for visualized data reasoning. A detailed description of each task is provided in Appendix~\ref{appendix:a.1}}
\label{tab:vis-tasks}
\end{table}


\section{Approach}
\label{Sec:3}

We propose VisDoT, a framework that enhances LVLMs' visual reasoning by emulating human visual decoding and incorporating LVLM-specialized inference. As illustrated in Figure~\ref{fig:framework_overview}, VisDoT comprises two components: (1) a perception-following dataset generation grounded in human perceptual priors, and (2) a DoT strategy that splits complex queries into sequential perception and logic sub-questions.
This section introduces the motivation and architecture of VisDoT, highlighting how each component contributes to complex visual reasoning. Section~\ref{Sec:3.1} describes the construction of perception-following questions based on structured visual representations. Section~\ref{Sec:3.2} presents the DoT strategy, which splits queries into perception–logic stages and produces aligned answers.


\begin{figure*}[!t]
    \centerline{\includegraphics[width=\textwidth]{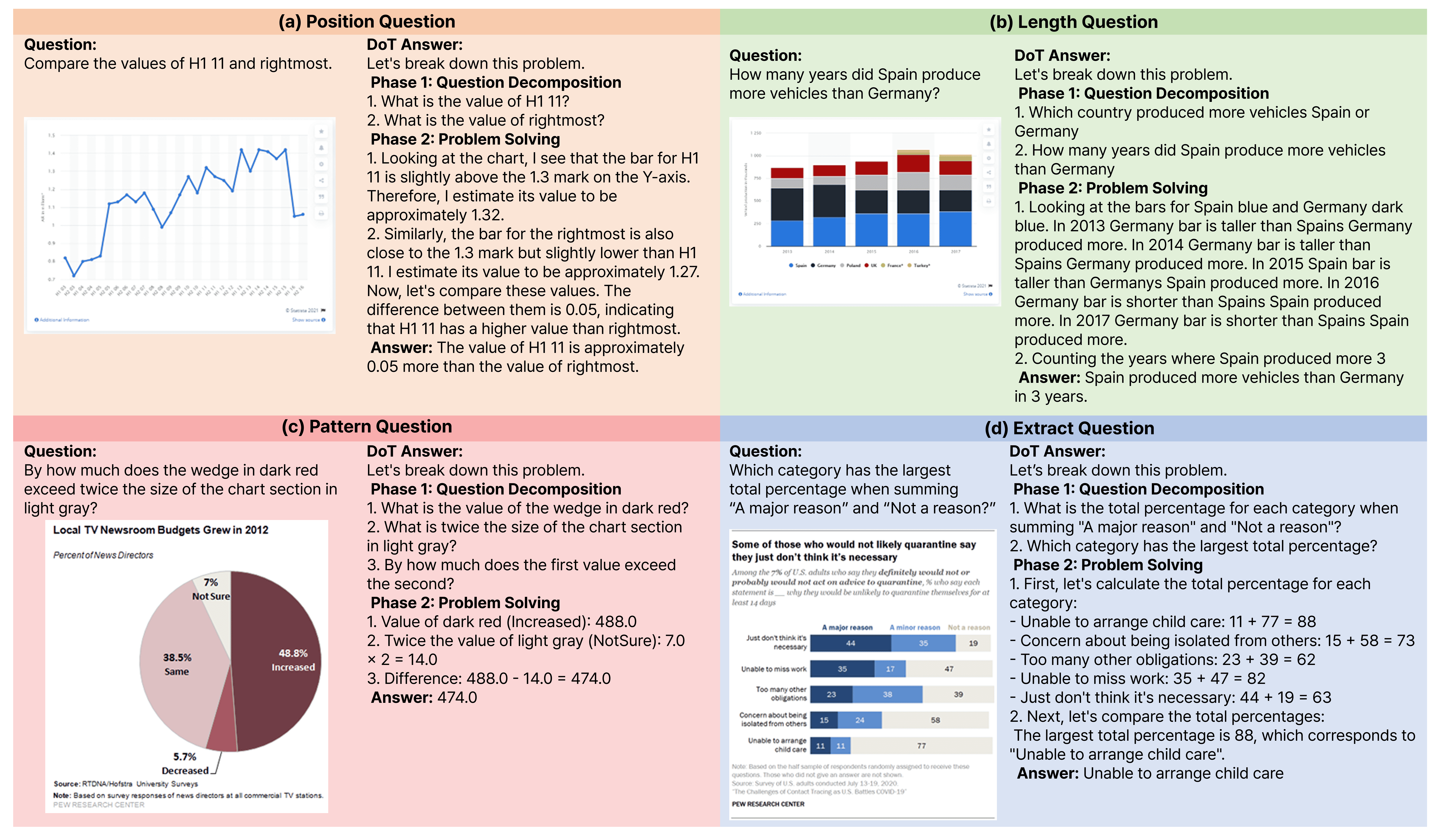}}
    \caption{Perception-following QA examples for the four task types in Table~\ref{tab:vis-tasks}. Each question is decomposed into perception and logic sub-questions using the DoT prompt, enabling structured and interpretable chart reasoning.}
    \label{fig:perceptual_task}
\end{figure*}


\subsection{Perception-following Question for LVLMs}
\label{Sec:3.1}
Human interpretation of visualized data involves decoding at the level of perceptual units, grounded in visual features such as position, length, or angle. Cognitive psychology research has shown that this process is hierarchically organized according to the accuracy and perceptual salience of these visual attributes~\cite{cleveland1984graphical}. However, existing LVLM training datasets predominantly consist of instruction-level QA pairs, making it difficult for models to learn or emulate such perceptual decoding processes~\cite{vogel2025refchartqa}.

\paragraph{Perceptual Task.} 
To lay the groundwork for modeling LVLMs that emulate human cognitive structures, we define a set of perceptual tasks (position, length, pattern, extract) grounded in human graphical cognition theory~\cite{cleveland1984graphical}. As shown in Table~\ref{tab:vis-tasks}, these tasks provide the basis for constructing training data that guides model attention to align with human visual information processing.

\paragraph{Perception-following Question Generation.}
Building on the defined perceptual tasks (Table~\ref{tab:vis-tasks}), we develop an automated question generation algorithm (\textcolor{red}{red arrow} in Figure~\ref{fig:framework_overview}) that constructs questions requiring visual perception-based reasoning. Depending on task characteristics, the generated questions may involve multi-object and multi-step inference. As illustrated in Figure~\ref{fig:perceptual_task}, they range from OCR-style queries (e.g., identifying "major" or "not", corresponding to sub-figure (d)) to those with explicit visual descriptors (e.g., "dark red" or "light gray", as shown in sub-figure (c)), and also include questions that require implicit visual interpretation based on chart axes or guide lines (sub-figures (a) and (b)). This approach enables the generation of questions that evaluate the model’s ability to interpret charts, align visual elements with linguistic references, and perform grounded reasoning. The full prompting template used for perception-following question generation is detailed in Appendix~\ref{appendix:d.1}.


\begin{figure}[!t]
    \centerline{\includegraphics[width=\columnwidth]{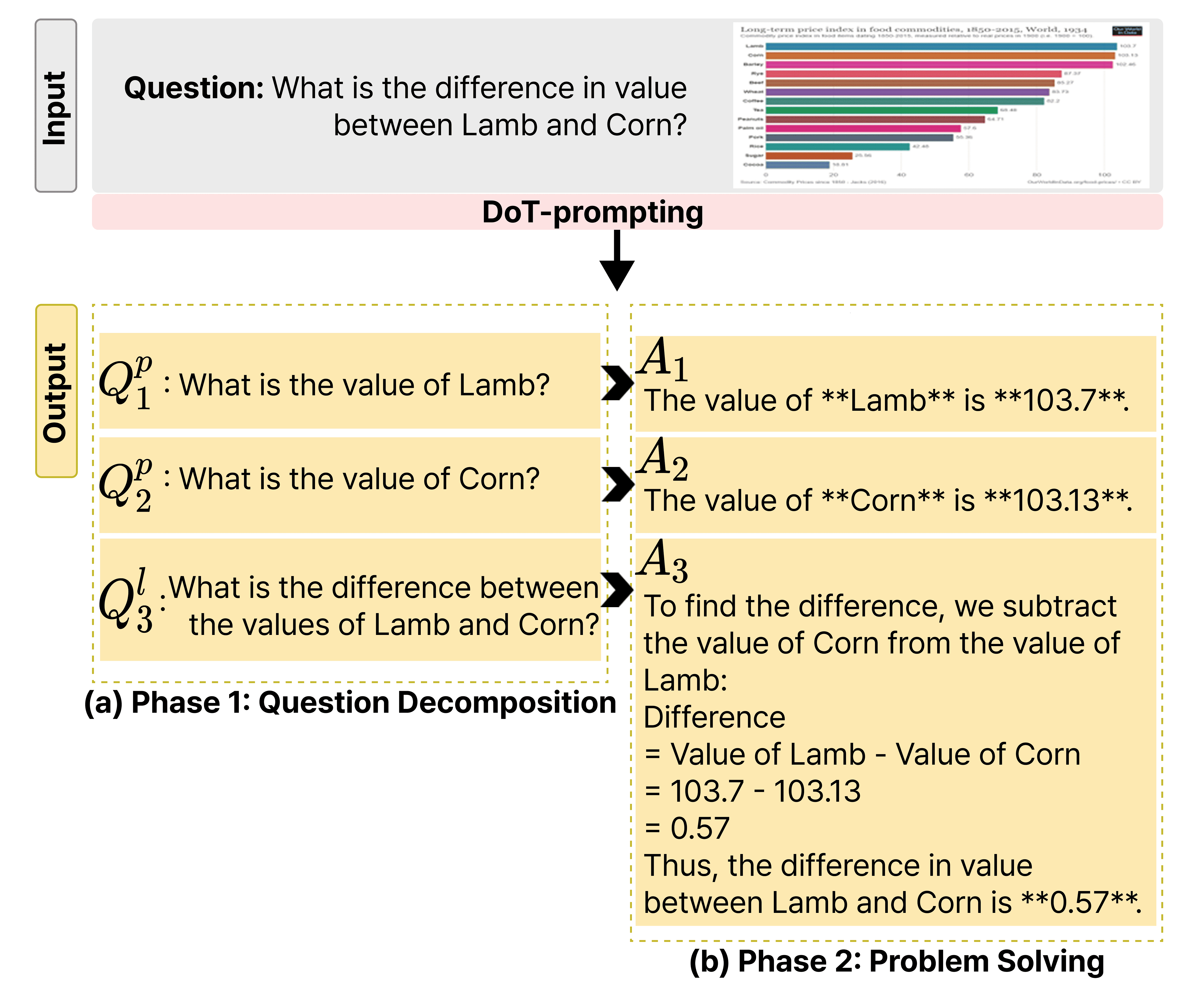}}
    \caption{The model is guided to decompose a complex visual question into perception and logic sub-questions (Question Decomposition) and generate intermediate reasoning steps sequentially (Problem Solving), enabling structured and interpretable visual inference.}
    \label{fig:dot_method} 
\end{figure}

\subsection{Decomposition-of-Thought Method}
\label{Sec:3.2}
Modular approaches that explicitly separate visual perception from logical reasoning have demonstrated strong performance and interpretability in complex VQA tasks, such as NS-VQA. In contrast, LVLMs typically employ language-centric reasoning strategies in multi-modal settings such as CoT. While these strategies are effective for symbolic tasks, they have shown limited effectiveness in scenarios that require grounded visual reasoning, as empirically demonstrated in Section~\ref{Sec:4.5}.

Motivated by this gap, we extend the principle of separating visual perception and logical reasoning to LVLMs and develop an LVLM-specialized visual reasoning strategy. In doing so, we redefine VQA as a compositional task and introduce the DoT method to guide LVLMs in solving these tasks. Finally, we generate answers by accumulating the phase-wise reasoning process enabled by the VisDoT framework as illustrated by the \textcolor{blue}{blue arrow} in Figure~\ref{fig:framework_overview}.

\paragraph{Visual Question Answering definition.}

Standard LVLM-based VQA aims to maximize the conditional likelihood of the answer $A$ given an image $I$ and a question $Q$, which can be formalized as the following objective~\cite{khan2023exploring}:
{\setlength{\abovedisplayskip}{4pt}
 \setlength{\belowdisplayskip}{4pt}
 \setlength{\abovedisplayshortskip}{2pt}
 \setlength{\belowdisplayshortskip}{2pt}
\begin{equation}
    \arg\max_{A} \; P(A \mid I, Q)
\label{eq:argmax}
\end{equation}
}
Following the separating visual perception and logical reasoning principle, we redefine VQA by formulating a question $Q$ as a compositional query that integrates visual perception and logical reasoning.

{\setlength{\abovedisplayskip}{4pt}
 \setlength{\belowdisplayskip}{4pt}
 \setlength{\abovedisplayshortskip}{2pt}
 \setlength{\belowdisplayshortskip}{2pt}
\begin{equation}\label{eq:decomp}
\begin{aligned}
P(A \mid I, Q) &= \sum_{\{Q_1, \ldots, Q_n\}} P(\{Q_1^p, \ldots, Q_n^l\} \mid Q) \notag \\
&\quad \cdot \prod_{i=1}^{n} P(A_i \mid I, Q_i, A_{<i})
\end{aligned}
\end{equation}
}
The compositional question $Q$ is represented as a set of sub-questions $Q_i$, each designed to solve a specific sub-task. Each sub-question $Q_i$ is classified as either a perception-oriented sub-question  $Q^p_i$ or a logic-oriented sub-question  $Q^l_i$. We do not impose constraints on the number of perception or logic-oriented sub-questions, which allows flexible decomposition of complex tasks into arbitrarily many sub-questions. For each sub-question  $Q_i$, the model generates an intermediate answer $A_i$ sequentially. The final answer $A_n$ is obtained by accumulating these intermediate answers. Each intermediate answer $A_i$ is generated based on the image $I$ and the preceding context $A_{<i}$, enabling context-aware multi-step reasoning. The first term $P(\{Q^p_1, Q^p_2, Q^l_3, \dots, Q^l_n\} \mid Q)$ represents the probability of decomposing the complex question $Q$ into an ordered sequence of perceptual $Q^p$ and logical $Q^l$ sub-questions~\cite{zhang2024far}. We assume that decompositions that prioritize perception first-order are generally more plausible~\cite{liu2025autonomousimaginationclosedloopdecomposition}. 
The second term, $\prod_{i=1}^n P(A_i \mid I, Q_i, A_{<i})$, models the likelihood of accurately answering each sub-question based on image-based and context-aware reasoning. 
Thus, the most reliable answers are obtained when the question is logically decomposed and each step is reasoned correctly.

\paragraph{DoT Prompt.}
To effectively perform chart reasoning as defined in Equation~\ref{eq:decomp}, we design a DoT prompt that explicitly guides LVLMs in structured answer generation. As illustrated in Figure~\ref{fig:dot_method}, the DoT prompt organizes the reasoning process into two phases.

\textbf{Phase 1: Question Decomposition.} The model decomposes a compositional question $Q$ into an ordered set of sub-questions consisting of perception-oriented $Q^p_i$ and logic-oriented $Q^l_i$ queries. The prompt explicitly enforces that perception-oriented sub-questions $Q^p_i$ are generated before logic-oriented $Q^l_i$ ones. As illustrated in Figure~\ref{fig:perceptual_task}(c), the prompt is designed to explicitly prioritize perception-oriented sub-questions (e.g., retrieving individual values or identifying visual properties) before logic-oriented ones (e.g., computing differences or performing comparisons). For example, questions 1 and 2 guide the model to extract visual information regarding chart components, while question 3 introduces a logical operation based on the previously grounded values.

\textbf{Phase 2: Problem Solving.} The model sequentially answers each sub-question $Q_i$ while explicitly accumulating intermediate reasoning steps. The resulting output (yellow box in Figure~\ref{fig:dot_method}) satisfies both components of Equation~\ref{eq:decomp}, thereby enabling structured and interpretable reasoning. The full prompt used in our framework is provided in Appendix~\ref{appendix:d}. To verify that the proposed DoT prompting is not limited to chart‑centric scenarios, we further evaluate its generalisation in Section~\ref{Sec:4.5} on two open‑domain VQA benchmarks that do \emph{not} involve charts. The same DoT template yields over CoT while markedly reducing visual‑grounding errors, confirming that the perception‑first decomposition strategy extends beyond chart reasoning to generic VQA tasks.

\paragraph{DoT-based Perception-following Answer Generation.}
As illustrated by the \textcolor{blue}{blue arrow} in Figure~\ref{fig:framework_overview}, the LVLM generates answers that include the DoT reasoning process guided by the DoT prompt. Figure~\ref{fig:perceptual_task} shows that the model produces reasoning trajectories tailored to the given image and question. As seen in Figure~\ref{fig:perceptual_task} (b), the model generates sub-questions that sequentially query multiple objects and constructs stepwise logic-oriented questions to handle complex reasoning. The answers generated in this manner explicitly reveal the fine-grained reasoning process, which enhances the interpretability of reasoning paths and improves adaptability to complex visual queries.
Finally, we train the LVLM on VisDoTQA, which is constructed using VisDoT, and conduct systematic evaluations against existing visual reasoning approaches on multiple benchmarks. The example instances from the generated dataset are shown in Appendix~\ref{appendix:e}.

\section{Experiment}
\label{Sec:4}

\subsection{Experimental Setup}
\label{Sec:4.1}

\begin{table}[t]
\small
\centering
\begin{tabularx}{\columnwidth}{@{}l>{\raggedleft\arraybackslash}X@{}}
\toprule
\textbf{Task Type} & \textbf{\# QA Pairs} \\
\midrule
\rowcolor{poscolor} Position & 66,617 \\
\rowcolor{lencolor} Length   & 31,614 \\
\rowcolor{patcolor} Pattern  & 61,268 \\
\rowcolor{extcolor} Extract  & 174,411 \\
\midrule
\textbf{Total} & \textbf{331,969} \\
\bottomrule
\end{tabularx}
\caption{Distribution of QA pairs across the four perceptual reasoning tasks in the VisDoTQA dataset.}
\label{tab:visdotqa_stats}
\end{table}

\paragraph{Data. }
We developed VisDoTQA using the VisDoT framework based on 16,167 chart images collected from Pew Research~\cite{pewresearch}, Statista~\cite{statista}, Our World in Data (OWID)~\cite{owid}, and the OECD~\cite{oecd}. The distribution of the dataset is presented in Table~\ref{tab:visdotqa_stats}. We fine-tune open-source models on VisDoTQA to validate the effectiveness of the VisDoT framework.
Question generation in Figure~\ref{fig:framework_overview} (\textcolor{red}{red arrow}) is performed using the closed-source GPT-4o~\cite{openai2024gpt4o}, while answer generation (\textcolor{blue}{blue arrow}) is conducted with the open-source LLaMA-3.2-90B~\cite{dubey2024llama}.

\paragraph{VisDoTQA test set. }
To prevent evaluation leakage, we hold out 609 charts that are never used in training or prompt development to construct VisDoTQA test set.
Questions are generated with GPT-4o~\cite{openai2024gpt4o}.
For answer labeling, we apply the DoT prompt to both \textbf{Llama-3.2-90B-Vision}~\cite{dubey2024llama} and \textbf{GPT-4o}~\cite{openai2024gpt4o}; if the two models produce identical final answers, we adopt it as the ground truth, otherwise a human verifies and edits the label.

\paragraph{Baseline.}
(i) As model baselines, we compare InternVL2.5~\cite{chen2024expanding}, Qwen2.5~\cite{bai2025qwen2}, and Gemma 3~\cite{team2025gemma}, a chart-specialized SOTA model ChartGemma~\cite{masry-etal-2025-chartgemma}, and closed-source models GPT-4o~\cite{openai2024gpt4o}, Gemini-Flash 2.0~\cite{comanici2025gemini}.
(ii) As methodology baselines on the same student backbone, we compare two chart distillation/data-synthesis pipelines with publicly released instruction data (ChartGemma~\cite{masry-etal-2025-chartgemma} and ECD~\cite{yang2025effective}). Full experimental settings are detailed in Appendix~\ref{appendix:a.2}.

\paragraph{Benchmarks.}
We evaluate our approach on three chart understanding benchmarks: ChartQA, ChartQAPro, and VisDoTQA. ChartQA and ChartQAPro span both human-authored and machine-generated questions, diverse chart types, and a wide range of reasoning tasks. VisDoTQA, constructed using our proposed framework, focuses on perceptual reasoning and serves as a challenging benchmark. Full dataset details are provided in Appendix~\ref{appendix:a.3}.
\paragraph{Evaluation metrics.}
We follow the official evaluation protocols for ChartQA and ChartQAPro, and use Relaxed Accuracy (RA) for VisDoTQA. Detailed evaluation metrics are provided in Appendix~\ref{appendix:a.3}.

\begin{table*}[t]
\scriptsize
\setlength{\tabcolsep}{2pt}
\centering
\begin{tabularx}{\textwidth}{@{}l|c|ccc|cccccc|XXXXX@{}}
\toprule
\textbf{Model} & \textbf{Param} & \multicolumn{3}{c|}{\textbf{ChartQA}} & \multicolumn{6}{c|}{\textbf{ChartQAPro}} & \multicolumn{5}{c}{\textbf{VisDoTQA}} \\
& & Mac. & Hum. & Avg. & Factoid & Multi. & Conv. & Check. & Hypo. & Avg. & Pos. & Len. & Pat. & Ext. & Avg. \\
\midrule
GPT-4o & -- & -- & -- & \textbf{85.7} & 35.76 & 46.72 & 34.75 & 45.49 & 28.91 & 37.67 & 74.57 & 55.00 & 36.33 & 57.03 & 57.14 \\
Gemini-Flash-2.0 & -- & 90.00 & 80.24 & 85.12 & \textbf{43.43} & \textbf{60.28} & \textbf{40.25} & \textbf{67.62} & 24.47 & \textbf{46.85} & 77.71 & \textbf{73.75} & 41.57 & 50.95 & 61.96 \\
\midrule
ChartGemma & 3B & 90.80 & 69.52 & 80.16 & 6.86 & 0.00 & 16.00 & 1.22 & 6.53 & 24.86 & 6.84 & 24.86 & 47.92 & 27.72 & 30.8 \\
InternVL & 2B & 83.52 & 64.80 & 74.16 & 13.86 & 10.74 & 14.02 & 45.90 & 18.92 & 17.81 & 44.29 & 54.17 & 19.10 & 17.87 & 34.20 \\
InternVL & 4B & 84.00 & 66.16 & 75.08 & 25.81 & 10.98 & 15.85 & 47.13 & 25.07 & 32.11 & 58.00 & 59.17 & 25.47 & 27.38 & 43.30 \\
VisDoT\_InternVL & 2B & 84.48 / 84.00$^{\dagger}$ & 72.72 & 78.60 / 78.36$^{\dagger}$ & 23.40 & 29.91 & 12.03 & 37.30 & 29.59 & 24.35 & 76.00 & 64.17 & 49.81 & 61.98 & 63.93 \\
VisDoT\_InternVL & 4B & 88.40 / \textbf{90.88}$^{\dagger}$ & \textbf{80.32} & 84.36 / 85.60$^{\dagger}$ & 32.75 & 47.20 & 15.38 & 53.28 & \textbf{40.82} & 34.54 & \textbf{87.14} & 69.58 & \textbf{67.42} & \textbf{77.95} & \textbf{76.52} \\
\bottomrule
\end{tabularx}
\caption{
Comparison of baseline and VisDoT-tuned models across ChartQA, ChartQAPro, and VisDoTQA. VisDoT models leverage perception-following datasets and DoT-based supervision. For entries reported as \textit{default / short-answer}$^{\dagger}$, the value after ``/'' is obtained under the short-answer (final-answer-only) setting. Abbreviations: Mac. = Machine, Hum. = Human, Avg. = Average, Multi. = Multi-choice, Conv. = Conversational, Check. = Fact-checking, Hypo. = Hypothetical, Pos. = Position, Len. = Length, Pat. = Pattern, Ext. = Extract.
}
\label{tab:main}
\end{table*}

\begin{table}[t]
\centering
\small
\resizebox{\columnwidth}{!}{%
\begin{tabular}{@{}l r c c c@{}}
\toprule
\textbf{Method} & \textbf{\# QA} & \textbf{ChartQA} & \textbf{ChartQAPro} & \textbf{VisDoTQA} \\
\midrule
ChartGemma & 163K & 83.96 & \textbf{36.61} & 26.79 \\
ECD & 320K & 75.24 & 17.24 & 58.30 \\
VisDoT (Ours) & 7.4K & \textbf{84.08} & 34.02 & \textbf{70.00} \\
\bottomrule
\end{tabular}%
}
\caption{Comparison of chart distillation/data-synthesis pipelines using publicly released instruction data: ChartGemma~\cite{masry-etal-2025-chartgemma} and ECD~\cite{yang2025effective}.
All methods are fine-tuned on the same backbone (InternVL2.5-4B) and evaluated on ChartQA, ChartQAPro, and VisDoTQA.
\# QA denotes the number of QA pairs used for fine-tuning; for VisDoT, we randomly sample 7.4K QA pairs for this comparison, while Table~\ref{tab:main} reports results using the full VisDoTQA training set (331,969 QA pairs).}
\label{tab:method_baseline}
\end{table}

\subsection{Main Result}
\label{Sec:4.2}
Table~\ref{tab:main} summarizes our results on ChartQA, ChartQAPro, and VisDoTQA. Fine-tuning InternVL with VisDoTQA (VisDoT\_InternVL) leads to consistent improvements over the InternVL baselines. On ChartQA, VisDoT\_InternVL‑2B and ‑4B achieve relative gains of +4.4\% and +9.3\%, reaching performance comparable to GPT‑4o and Gemini‑Flash‑2.0. The gains are more pronounced on the Human split (+7.9\% and +14.2\%). When constrained to short-answer output, VisDoT\_InternVL‑4B achieves 90.88\% (+6.9\%), establishing a new state of the art. Appendix~\ref{appendix:c.1} further illustrates the importance of removing unnecessary reasoning steps in OCR-style questions.
On ChartQAPro, VisDoT\_InternVL‑2B achieves 24.35\% (+6.5\%), while the 4B variant reaches 34.54\% (+2.4\%). Notable gains are observed in fact-checking (53.28\%, +7.8\%), multi-choice (47.20\%), and hypothesis reasoning (+11.9\% and +16.4\%), surpassing GPT‑4o. Despite limited training image diversity, the models generalize well to 157 unseen domains and unfamiliar chart layouts, demonstrating the effectiveness of perception-oriented tuning for domain generalization. As shown in Appendix~\ref{appendix:c.2}, baseline models often fail on charts with novel structures or multi-object configurations, whereas VisDoT\_InternVL successfully handles these cases via DoT-based decomposition.
Finally, on VisDoTQA, VisDoT\_InternVL‑4B improves the average accuracy from 43.30\% to 76.52\% (+33.2\%), outperforming GPT‑4o (+19.4\%) and Gemini (+14.6\%). The most significant gains appear in the Position (+29.1\%), Length (+10.4\%), Pattern (+42.0\%) tasks, and Extract (+49.6\%). These findings confirm that DoT-based perception–logic decomposition substantially strengthens visual grounding and compositional reasoning, enabling mid-sized LVLMs to match or exceed large closed models on both chart-specific and general visual question answering benchmarks. Additional model variants presented in Appendix~\ref{appendix:b.1} exhibit consistent trends.
\paragraph{Comparison with chart distillation/data-synthesis pipelines.} In addition to model baselines, we compare our approach with two representative chart distillation/data-synthesis pipelines that release instruction data publicly: ChartGemma~\cite{masry-etal-2025-chartgemma} and ECD~\cite{yang2025effective}. Table~\ref{tab:method_baseline} summarizes the results under the same student backbone. Despite using substantially fewer QA pairs (7.4K vs. 163K/320K), VisDoT achieves the best performance on ChartQA (84.08\%) and VisDoTQA (70.00\%). Notably, VisDoTQA shows a large margin over ChartGemma (+43.21\%) and ECD (+11.70\%), suggesting that perception-following supervision paired with DoT-based decomposition is significantly more effective for perceptual reasoning than scaling synthetic instruction data alone. On ChartQAPro, VisDoT remains competitive (34.02\%), outperforming ECD by a wide margin (+16.78\%) while approaching ChartGemma (36.61\%). These results indicate that VisDoT yields strong generalization across chart understanding benchmarks, particularly on perceptual-heavy tasks, while requiring orders of magnitude fewer fine-tuning QA pairs.

\begin{table*}[t]
\small
\centering
\begin{tabularx}{\textwidth}{@{}l|ccc|ccccc@{}}
\toprule
\textbf{Method} & \multicolumn{3}{c|}{\textbf{ChartQA}} & \multicolumn{5}{c}{\textbf{VisDoTQA}} \\
& Machine & Human & Avg. & Extract & Position & Length & Pattern & Avg. \\
\midrule
Base  & 84.00 & 66.16 & 75.08 & 58.00 & 59.17 & 25.47 & 27.38 & 43.30 \\
CoT-trained model & 73.96 / 87.68$^{\dagger}$ & 63.92 & 68.94 / 75.8$^{\dagger}$ & 57.14 & 52.50 & 30.34 & 56.27 & 49.55 \\
Perceptual-trained model & 91.20 & 72.56 & 81.88 & 65.43 & 62.92 & 22.10 & 28.14 & 45.80 \\
DoT-trained model & 89.68 / \textbf{91.28}$^{\dagger}$ & 76.24 & 82.96 / 83.76$^{\dagger}$ & 68.86 & \textbf{66.25} & 51.31 & 66.54 & 63.57 \\
Perceptual + DoT (Ours) & 83.44 / 91.20$^{\dagger}$ & \textbf{76.96} & 80.20 / \textbf{84.08}$^{\dagger}$ & \textbf{78.57} & 65.83 & \textbf{61.42} & \textbf{71.10} & \textbf{70.00} \\
\bottomrule
\end{tabularx}
\caption{Performance comparison of training strategies on ChartQA and VisDoTQA using InternVL2.5-4B as the base model. $^{\dagger}$ denotes short-answer evaluation (final-answer-only).}
\label{tab:ablation}
\end{table*}

\subsection{Ablation Study: Effect of Perceptual Alignment and Thought Decomposition}
\label{Sec:4.3}

Table~\ref{tab:ablation} compares four training regimes on ChartQA and VisDoTQA, each using 7.4K matched samples (see Appendix~\ref{appendix:a.4} for details). CoT reduces average accuracy on ChartQA, particularly on machine‑generated questions, but regains a small margin when answers are forced into short form. On VisDoTQA, the same template raises accuracy, confirming its value for logic-only tasks, yet it also reveals sensitivity to question difficulty (Section \ref{Sec:4.5}). Perceptual yields the largest gains on ChartQA and moderate gains on VisDoTQA, indicating strong visual alignment but limited reasoning capacity. The DoT improves both datasets and remains stable across shifts in difficulty. On the ChartQA machine data, it avoids the -10.0\% drop seen with CoT and instead scores +7.9\% (free form) and +8.7\% (short form). Perceptual + DoT maintains the Perceptual gains while adding DoT robustness, giving +9.0\% on ChartQA and +20.3\% on VisDoTQA, a 3.2× larger boost than CoT on the latter.
These results show that visual alignment and decomposition are complementary. Normalising answers to short form further sharpens performance, and the combined Perceptual + DoT strategy delivers reliable improvements across both OCR‑style extraction and compositional reasoning tasks.

\begin{table}[t]
\small
\centering
\begin{tabularx}{\columnwidth}{@{}lccccc@{}}
\toprule
\textbf{Model} & \textbf{Pos.} & \textbf{Len.} & \textbf{Pat.} & \textbf{Ext.} & \textbf{Avg.} \\
\midrule
\textbf{All} & {\scriptsize 83.43\%} & {\scriptsize 63.33\%} & {\scriptsize 63.30\%} & {\scriptsize 74.90\%} & {\scriptsize \textbf{72.32\%}} \\
\textbf{w/o Position} & {\scriptsize 73.43\%} & {\scriptsize 66.25\%} & {\scriptsize 63.30\%} & {\scriptsize 72.24\%} & {\scriptsize \textbf{69.20\%}} \\
\textbf{w/o Length} & {\scriptsize 85.14\%} & {\scriptsize 65.83\%} & {\scriptsize 64.04\%} & {\scriptsize 70.34\%} & {\scriptsize \textbf{72.50\%}} \\
\textbf{w/o Pattern} & {\scriptsize 84.00\%} & {\scriptsize 64.58\%} & {\scriptsize 57.68\%} & {\scriptsize 69.96\%} & {\scriptsize \textbf{70.27\%}} \\
\bottomrule
\end{tabularx}
\caption{Performance across four perceptual tasks under ablation settings. For brevity, we denote Position as \textbf{Pos.}, Length as \textbf{Len.}, Pattern as \textbf{Pat.}, Extract as \textbf{Ext.},  and Average as \textbf{Avg.}.}
\label{tab:perceptual_task}
\end{table}

\begin{table}[t]
\small
\centering
\begin{tabularx}{\columnwidth}{@{}l>{\centering\arraybackslash}X>{\centering\arraybackslash}X>{\centering\arraybackslash}X@{}}
\toprule
\textbf{Dataset} & \textbf{Direct} & \textbf{CoT} & \textbf{Ours (DoT)} \\
\midrule
POPE & 84.29\% & 84.64\% & \textbf{86.07\%} \\
MMMU & 17.7\%  & 35.5\%  & \textbf{37.7\%} \\
\bottomrule
\end{tabularx}
\caption{Zero-shot performance comparison on POPE and MMMU benchmarks using Direct prompting, CoT, and our proposed DoT strategy.}
\label{tab:cross_domain}
\end{table}

\subsection{Ablation Study: Role of Individual Perceptual Task}
\label{Sec:4.4}
Table~\ref{tab:perceptual_task} reports the results of fine-tuning VisDoT while sequentially removing each perceptual task from the training data. For fair comparison, 30K samples per task were randomly selected from VisDoTQA. Removing Position data produces the steepest decline: the average drops from 72.32\% to 69.20\% (- 3.12\%), Position from 83.43\% to 73.43\% (- 10.00\%), and Extract from 74.90\% to 72.24\% (- 2.66\%). Position cues, therefore, support both spatial alignment and downstream extraction. Removing the Length data has a negligible or positive impact. w/o‑Length scores 65.83\% on Length tests (+2.50\%), and w/o‑Position scores 66.25\% (+2.92\%). On Position tests, w/o‑Length attains 85.14\% (+1.71\%). Position thus substitutes for Length and reveals feature competition consistent with cognitive findings. More data alleviate this competition (Table \ref{tab:main}).

Overall, Position is critical for stable accuracy and extraction, Length is partly redundant, and Pattern governs categorical discrimination. Perception‑following fine‑tuning with ample samples remains essential for reliable reasoning on visualized data.

\subsection{Cross-Domain Evaluation of DoT Prompt's Robustness}
\label{Sec:4.5}

Table~\ref{tab:cross_domain} evaluates DoT under zero‑shot transfer using InternVL2.5‑78B on POPE and MMMU. POPE targets natural object images, whereas MMMU spans six academic domains and 30 visual formats. DoT exceeds both Direct and CoT baselines: +1.43\% on POPE and +2.2\% over CoT on MMMU. These findings suggest that DoT is not a simple variant of CoT but a structured reasoning strategy grounded in the separation of perception and logic.
The consistent improvements across natural image QA and diverse visual domains indicate that DoT has strong potential for domain-invariant transfer. Furthermore, DoT’s explicit separation of perceptual and logical sub-questions enables stable reasoning over decomposed sub-queries, supporting its role as a general-purpose reasoning strategy for LVLMs. 
Moreover, DoT satisfies both components of the structured reasoning formulation in Equation~\ref{eq:decomp}, enabling more robust and generalizable visual reasoning beyond chart-centric settings. Detailed experimental results are provided in Appendix~\ref{appendix:b.2}.

\section{Conclusion}
\label{Sec:5}
This work introduced VisDoT, a framework for visual question answering on data visualizations. Grounded in graphic perception theory, we formalized four perception‑following task families (Position, Length, Pattern, and Extract) and designed a DoT prompting strategy that first performs perceptual grounding and then logical reasoning. VisDoT enables robust visual grounding even with compact LVLMs and produces interpretable step‑wise reasoning traces. We also release a perception‑following dataset that embodies these design principles.
Experiments across diverse visualization types show that current LVLMs still face substantial challenges in understanding visualizations and that grounding accuracy is a primary bottleneck in visual reasoning. VisDoT mitigates this limitation by explicitly integrating grounding with reasoning through the DoT prompt. Future work will extend this approach to other grounding‑intensive domains and will leverage DoT reasoning traces to analyze whether LVLM failures stem from deficits in perception or reasoning.

\section*{Limitations}
While the current study primarily evaluated the framework using raster chart formats (e.g., PNG, JPEG), future work could further investigate its applicability to vector-based formats (e.g., SVG, PDF) that preserve object boundaries, textual metadata, and hierarchical group information, potentially enabling richer grounding capabilities. Extending the VisDoT framework to more complex visualization layouts, such as multi-panel dashboards with interactive elements, may provide additional insights. Moreover, although VisDoT focuses on perception tuning for enhanced visual grounding, incorporating instruction tuning could better address real-world queries involving higher-order discourse structures (e.g., comparison, explanation, prediction). Future research could explore hybrid training strategies that jointly optimize visual token recognition and discourse-aware response quality, possibly through curriculum-based approaches interleaving perception-first and instruction-rich samples. Finally, prompt design for DoT remains an open area for refinement, and further investigations are needed to identify strategies that can maximize the effectiveness and generalizability of the framework.

\section*{Acknowledgments}
This research was supported by the MSIT(Ministry of Science and ICT), Korea, under the ITRC(Information Technology Research Center) support program(IITP-2026-RS-2020-II201789), and the Artificial Intelligence Convergence Innovation Human Resources Development(IITP-2026-RS-2023-00254592) supervised by the IITP(Institute for Information \& Communications Technology Planning \& Evaluation).

\appendix

\section{Detailed Descriptions}
\label{appendix:a}

\subsection{Perceptual Task Details}
\label{appendix:a.1}
As briefly introduced in Section~\ref{Sec:3.1}, we define four perceptual tasks for visualized data interpretation: Position, Length, Pattern and Extract. These tasks are based on foundational principles of graphical perception~\cite{cleveland1984graphical}. This section provides a more detailed explanation of the rationale behind selecting these perceptual tasks. 

Based on the prior work of Cleveland and McGill~\cite{cleveland1984graphical}, we select three core perceptual tasks—\textbf{Position}, \textbf{Length}, and \textbf{Pattern}—as foundational operations for visual reasoning. Their study identified ten elementary perceptual tasks essential for chart interpretation and ranked them according to their perceptual accuracy. Among these, Position demonstrated the highest interpretive accuracy in human studies and is therefore included as a primary visual task. Length belongs to the second-highest accuracy tier and, according to Stevens' power law~\cite{stevens2017psychophysics}, is one of the least distorted visual encodings.
In contrast, other candidates, such as angle, area, and volume, were excluded due to their known susceptibility to perceptual bias and overestimation. The task pattern—while not suitable for precise quantitative estimation—is included for its effectiveness in categorical differentiation, as emphasized by~\cite{cleveland1984graphical}. The \textbf{Extract} task refers to directly retrieving explicitly labeled numerical values from the visual input. For instance, to answer the question “What is the difference in GDP between the United States and Canada?”, the model must directly extract the GDP values of both countries from the chart.

In summary, we define the following four perceptual tasks as the fundamental components for visual reasoning over visualized data: Position, Length, Pattern, and Extract.

\begin{table}[t]
\small
\centering
\begin{tabularx}{\columnwidth}{@{}l>{\centering\arraybackslash}X>{\centering\arraybackslash}X>{\centering\arraybackslash}X>{\centering\arraybackslash}X@{}}
\toprule
\textbf{Model} & \textbf{Epoch} & \textbf{Learning Rate} & \textbf{Batch Size} & \textbf{Hours} \\
\midrule
InternVL-2B     & 1 & 4e-5 & 16 & 10 \\
InternVL-4B     & 1 & 4e-5 & 10 & 12 \\
Gemma3-4B       & 1 & 2e-4 & 4  & 20 \\
Qwen2.5-VL-3B   & 1 & 2e-4 & 64 & 29 \\
\bottomrule
\end{tabularx}
\caption{Training configurations used for different LVLM backbones, including number of epochs, learning rate, batch size, and training time.}
\label{tab:training_config}
\end{table}

\subsection{Training Configurations}
\label{appendix:a.2}
We fine-tuned four LVLMs under different training strategies, including parameter-efficient LoRA-based tuning and full fine-tuning. The specific configuration for each model is detailed below, and its corresponding hyper-parameters are summarized in Table~\ref{tab:training_config}. InternVL2.5-2B and InternVL2.5-4B were fine-tuned using the LoRA method (rank = 16), applied exclusively to the language decoder. Both the vision encoder and the multi-modal projector were kept frozen during training. Gemma3-4B was fine-tuned using the Unsloth framework in a parameter-efficient manner. The vision encoder was frozen, and LoRA adapters (rank = 8, $\alpha$ = 16) were applied to the language model. Qwen2.5-VL-3B was fine-tuned using a full fine-tuning strategy with the exception of the vision encoder, which remained frozen. Both the multi-modal MLP and the language model components were updated during training to adapt the model to downstream tasks. We used six A6000 GPUs, and the total training took approximately 71 hours.

\subsection{Benchmarks, Metrics and Models Details}
\label{appendix:a.3}

\paragraph{Benchmarks}
We evaluate our approach on three chart understanding benchmarks. ChartQA consists of two subsets: ChartQA-Human, which contains human-authored questions, and ChartQA-Machine, which includes questions generated from chart summaries. ChartQAPro is built from unseen charts collected across 157 domains, covering not only simple bar and line charts but also complex visual structures such as multi-chart layouts, stacked bar charts, dashboards, and infographics. It includes five reasoning types: factoid, conversational, multiple-choice, hypothetical, and fact-checking. We further evaluate on VisDoTQA, which is constructed using the VisDoT framework with images that are not used in training. VisDoTQA comprises questions incorporating four types of perceptual information and requires multi-object perception, grounding, and advanced reasoning, providing a challenging benchmark for comprehensive chart understanding.

\paragraph{Evaluation Metrics}
For ChartQA, we adopt the Relaxed Accuracy (RA) metric for evaluation. For ChartQAPro, evaluation is performed as follows. Numerical answers are considered correct if they fall within a ±5\% error margin. For year values, an exact match (EM) is required. Textual answers are evaluated using the average normalized levenshtein similarity (ANLS~\cite{biten2019scene}) score. Multiple-choice (MCQ) and fact-checking questions are assessed using the EM criterion. For VisDoTQA, we employ the RA metric for evaluating answers.

\paragraph{Models}
We fine-tuned Gemma3-4B (Apache 2.0 license) using the Unsloth framework (Apache 2.0 license), applying LoRA (rank = 8, $\alpha$ = 16) to the text decoder. Qwen2.5-VL-3B (Apache 2.0 license) was fully fine-tuned except for the frozen vision encoder, and InternVL2.5-2B/4B (MIT license) models were fine-tuned using LoRA (rank = 16) applied only to the language decoder. For baseline inference, we used the Ollama platform (MIT license). All models were trained and evaluated on the ChartQA and ChartQAPRO datasets (MIT license). We also referred to the ChartGemma implementation (GPL-3.0 license) during preliminary experiments.

\subsection{Detailed Description of the Ablation Study}
\label{appendix:a.4}
To evaluate the effectiveness of our proposed Perceptual tasks and decomposition-of-thought (DoT) strategy in visualized data reasoning, we design five training configurations using the same image–question dataset. For each of the 7,400 image–question pairs, responses are generated under the following distinct paradigms:
\begin{itemize}
    \item \textbf{Base:} Direct answers are generated without any intermediate reasoning structure or prompt engineering, serving as the most naive baseline.
    \item \textbf{CoT:} Responses are generated by reasoning through a step-by-step process before producing the final answer.
    \item \textbf{Perceptual:} Each sample includes explicit visual cues—such as spatial location, color, and order—prior to answering. While the output format remains consistent with Base, the model is guided to develop a perception-aligned reasoning mechanism.
    \item \textbf{DoT:} Unlike CoT, this configuration decomposes complex questions into a series of sub-questions. The model sequentially answers each sub-question to arrive at the final answer.
    \item \textbf{Perceptual + DoT (Ours):} This is the most structured setting, combining both perceptual grounding and decompositional reasoning. Each instance involves a two-stage process: the Question Phase decomposes the original query into perceptual and cognitive sub-questions, and the Answer Phase sequentially generates intermediate answers leading to the final response.
\end{itemize}
These five configurations form the basis of the comparative experiments shown in Table~\ref{tab:ablation}, providing a comprehensive analysis of the impact of our VisDoT framework on the performance of LVLMs.

\section{Additional Results and Analysis}
\label{appendix:b}

\begin{table*}[t]
\scriptsize
\setlength{\tabcolsep}{2pt}
\centering
\begin{tabularx}{\textwidth}{@{}l|X|XXX|XXXXXX|XXXXX@{}}
\toprule
\textbf{Model} & \textbf{Param} & \multicolumn{3}{c|}{\textbf{ChartQA}} & \multicolumn{6}{c|}{\textbf{ChartQAPro}} & \multicolumn{5}{c}{\textbf{VisDoTQA}} \\
& & Mac. & Hum. & Avg. & Factoid & Multi. & Conv. & Check. & Hypo. & Avg. & Pos. & Len. & Pat. & Ext. & Avg. \\
\midrule
Gemma (CoT) & 4B & -- & 58.24 & -- & 11.01 & 24.77 & -- & 47.13 & 12.24 & -- & 46.86 & 38.75 & 17.98 & 43.35 & 37.41 \\
Qwen (CoT) & 3B & -- & 67.12 & -- & 24.46 & 32.24 & 5.40 & 45.49 & 33.97 & -- & 67.43 & 54.58 & 48.69 & 65.78 & 59.82 \\
\midrule
VisDoT\_Gemma & 4B & 84.16 & 77.04 & 80.60 & 23.40 & 28.04 & -- & \textbf{54.10} & 38.78 & -- & \textbf{79.14} & \textbf{58.75} & 59.18 & 70.72 & 67.32 \\
VisDoT\_Qwen & 3B & \textbf{86.96} & \textbf{79.60} & \textbf{83.28} & \textbf{26.46} & \textbf{37.38} & -- & 43.44 & \textbf{40.82} & -- & 77.14 & 55.42 & \textbf{59.55} & \textbf{73.00} & \textbf{68.04} \\
\bottomrule
\end{tabularx}
\caption{
Comparison of CoT and VisDoT-tuned models across ChartQA, ChartQAPro, and VisDoTQA. VisDoT models leverage perception-following datasets and DoT-based supervision. Gray cells (if any) denote short-answer performance. Abbreviations: Mac. = Machine, Hum. = Human, Avg. = Average, Multi. = Multi-choice, Conv. = Conversational, Check. = Fact-checking, Hypo. = Hypothetical, Pos. = Position, Len. = Length, Pat. = Pattern, Ext. = Extract.
}
\label{tab:main_cot_visdot}
\end{table*}

\subsection{Additional Results using Other Models}
\label{appendix:b.1}
In addition to the experimental results presented in Section~\ref{Sec:4.2}, we provide supplementary results using alternative model backbones. Our results are shown in Table~\ref{tab:main_cot_visdot}. These additional experiments consistently corroborate the findings reported in the main text, further validating the robustness and generalizability of our conclusions.

\subsection{Detailed Results for POPE, MMMU}
\label{appendix:b.2}
To validate the generalizability of our proposed DoT prompting strategy, we evaluate it on two external benchmarks in a zero-shot setting: POPE~\cite{li2023evaluating}, which targets hallucination detection in natural images, and MMMU~\cite{yue2024mmmu}, which spans diverse academic visual reasoning tasks.

We independently sample \textbf{280 questions each} from POPE and MMMU to ensure consistent and controlled comparisons across all prompting strategies.

\vspace{1mm}
\noindent\textbf{Quantitative Results.}
As summarized in Table~6, DoT consistently outperforms both Direct and CoT prompting baselines across both benchmarks:

\begin{itemize}
  \item \textbf{POPE}: DoT achieves \textbf{86.07\%} accuracy, outperforming CoT (84.64\%) and Direct (84.29\%) by \textbf{+1.43\%p} and \textbf{+1.78\%p}, respectively.
  \item \textbf{MMMU}: DoT reaches \textbf{37.7\%}, improving over CoT (35.5\%) by \textbf{+2.2\%p} and over Direct (17.7\%) by a substantial \textbf{+20.0\%p}.
\end{itemize}

These results demonstrate that DoT is more robust across both binary judgment tasks (POPE) and multi-format reasoning benchmarks (MMMU).

\vspace{1mm}
\noindent\textbf{Why is DoT More Effective?}
Unlike CoT, which decomposes every question into multi-step reasoning regardless of complexity, DoT dynamically determines when reasoning is necessary based on perceptual cues. This perception-first principle allows DoT to avoid over-decomposition and reduce hallucinations, especially when visual evidence alone suffices.

\begin{itemize}
    \item In \textbf{POPE}, CoT sometimes hallucinates object presence due to speculative reasoning (e.g., ``There might be a stool beside the plant''). DoT, in contrast, first verifies perceptual absence (e.g., ``No object resembling a chair is present'') before concluding.
    \item In \textbf{MMMU}, CoT often misinterprets axis semantics or misaligns chart elements due to rigid, uniform reasoning chains. DoT first inspects structural layout (e.g., axis labels, bar heights), and then performs reasoning grounded in perceptual understanding.
\end{itemize}

\vspace{1mm}
\noindent\textbf{POPE Example:}
\begin{itemize}
  \item \textbf{Question:} Is there a chair in the image?
  \item \textbf{Answer:} No
  \item \textbf{Direct:} No chair visible. (Correct)
  \item \textbf{CoT:} There is a stool beside the plant. (Incorrect – hallucination)
  \item \textbf{DoT:} No object resembling a chair (with back and legs) is present. (Correct)
\end{itemize}

\vspace{1mm}
\noindent\textbf{MMMU Example:}
\begin{itemize}
  \item \textbf{Question:} What circumstances lead to the typical growth form of the trees shown in the image?
  \item \textbf{Answer:} D
  \item \textbf{Direct:} r (Incorrect – uninformative)
  \item \textbf{CoT:} Focuses on bark pattern but overlooks canopy structure. (Incorrect)
  \item \textbf{DoT:} Inspects overall morphology and spacing, then selects D. (Correct)
\end{itemize}

These examples demonstrate that DoT not only improves accuracy but also enhances interpretability and factual consistency over existing prompting strategies.

\section{Case Studies}
\label{appendix:c}


\begin{figure*}[!t]
    \centerline{\includegraphics[width=\textwidth]{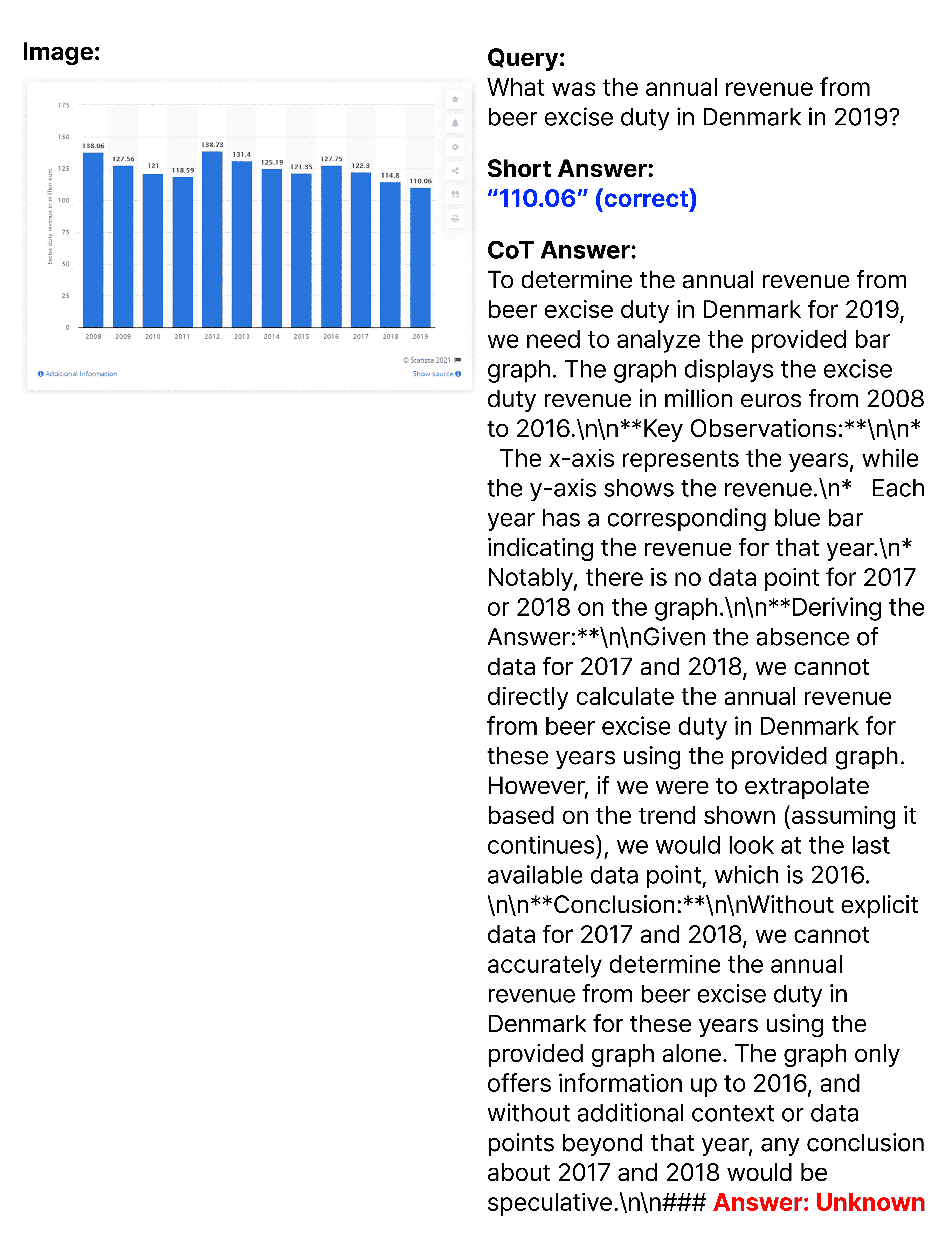}}
    \caption{Short answer case 1}
    \label{fig:short_answer_case1}
\end{figure*}


\begin{figure*}[!t]
    \centerline{\includegraphics[width=\textwidth]{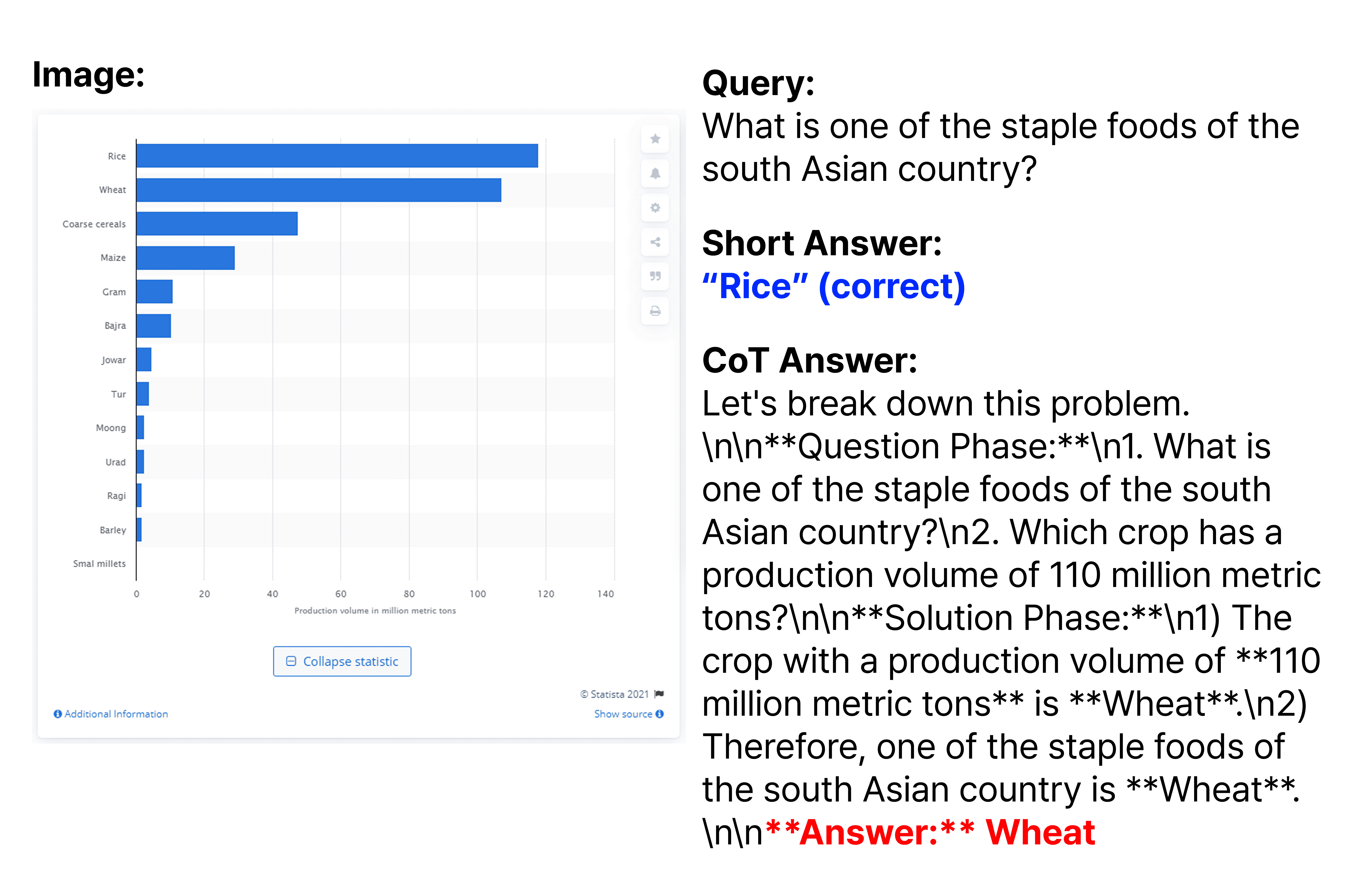}}
    \caption{Short answer case 2}
    \label{fig:short_answer_case2}
\end{figure*}

\subsection{How Unnecessary Reasoning Steps Degrade Performance?}
\label{appendix:c.1}
In OCR-style tasks involving relatively simple questions, directly providing short answers proves to be more effective than generating extended reasoning traces. As shown in Figures~\ref{fig:short_answer_case1} and \ref{fig:short_answer_case2}, unnecessary decomposition or step-by-step reasoning may introduce errors or distract the model from the correct answer, ultimately degrading performance. These findings highlight that, for low-complexity visual queries, minimal reasoning with concise output formatting can lead to superior accuracy.


\begin{figure*}[!t]
    \centerline{\includegraphics[width=\textwidth]{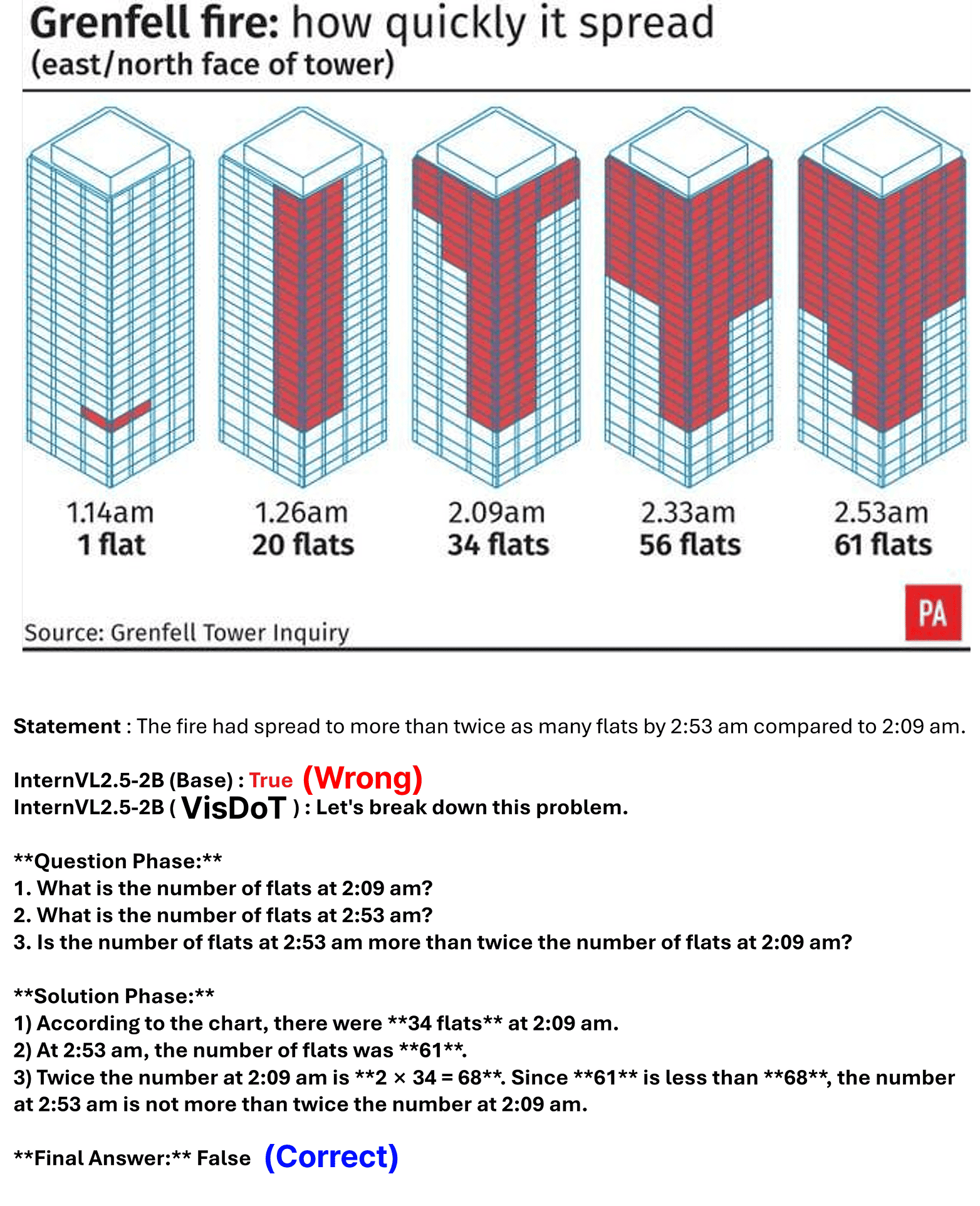}}
    \caption{DoT adv example}
    \label{fig:dot_adv_example}
\end{figure*}

\subsection{Advantages of DoT Prompting}
\label{appendix:c.2}
Base models often fail to reason over visualized data involving unseen visual structures or multi-object references. In contrast, models trained with VisDoTQA effectively utilize the DoT-style question decomposition strategy to derive the correct answers. A representative example is shown in Figure~\ref{fig:dot_adv_example}.

\begin{figure*}[!t]
    \centering
    \includegraphics[width=\textwidth]{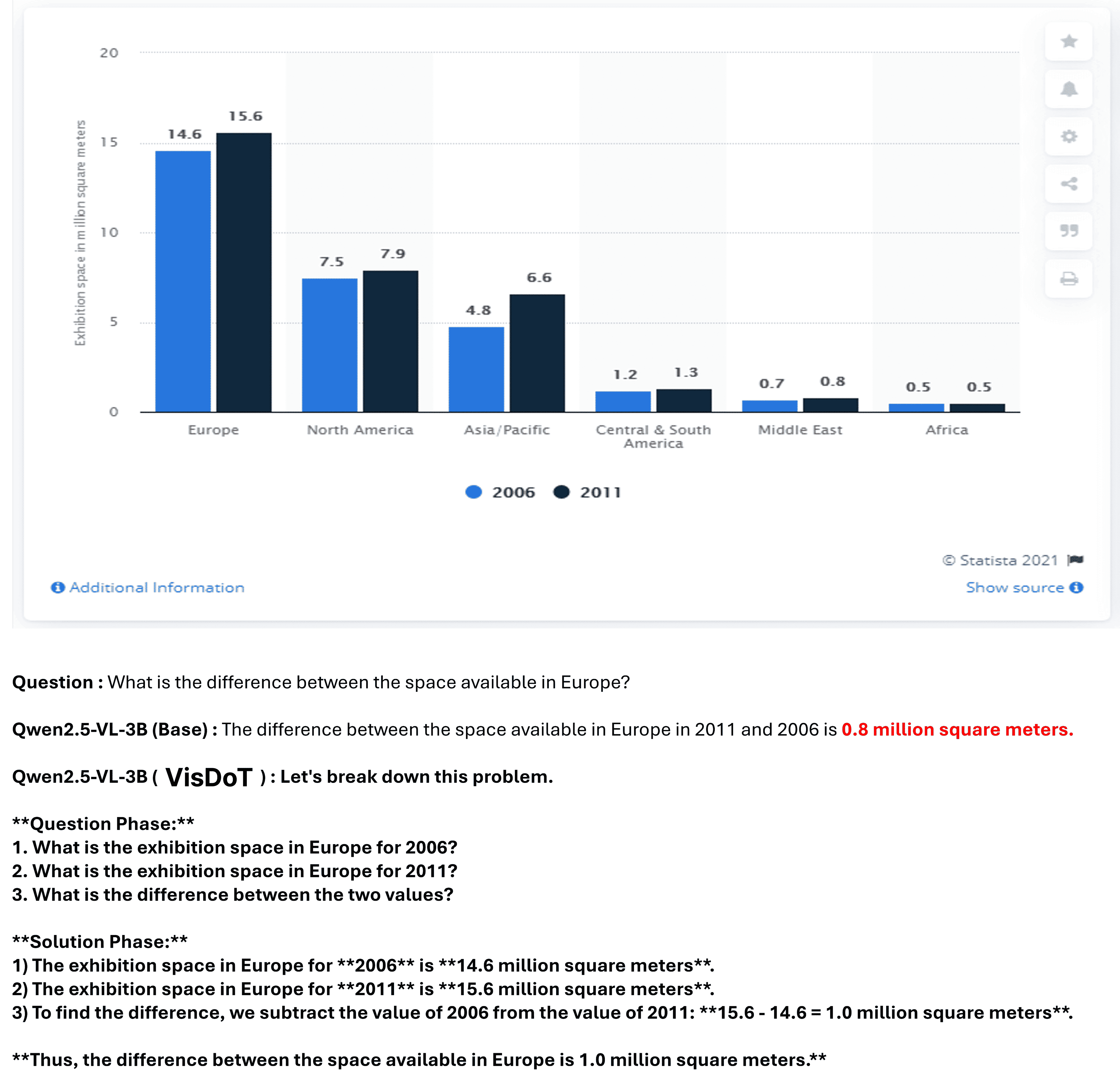}
    \caption{ChartQA Human task 1}
    \label{fig:chartqa1}
\end{figure*}

\begin{figure*}[!t]
    \centering
    \includegraphics[width=\textwidth]{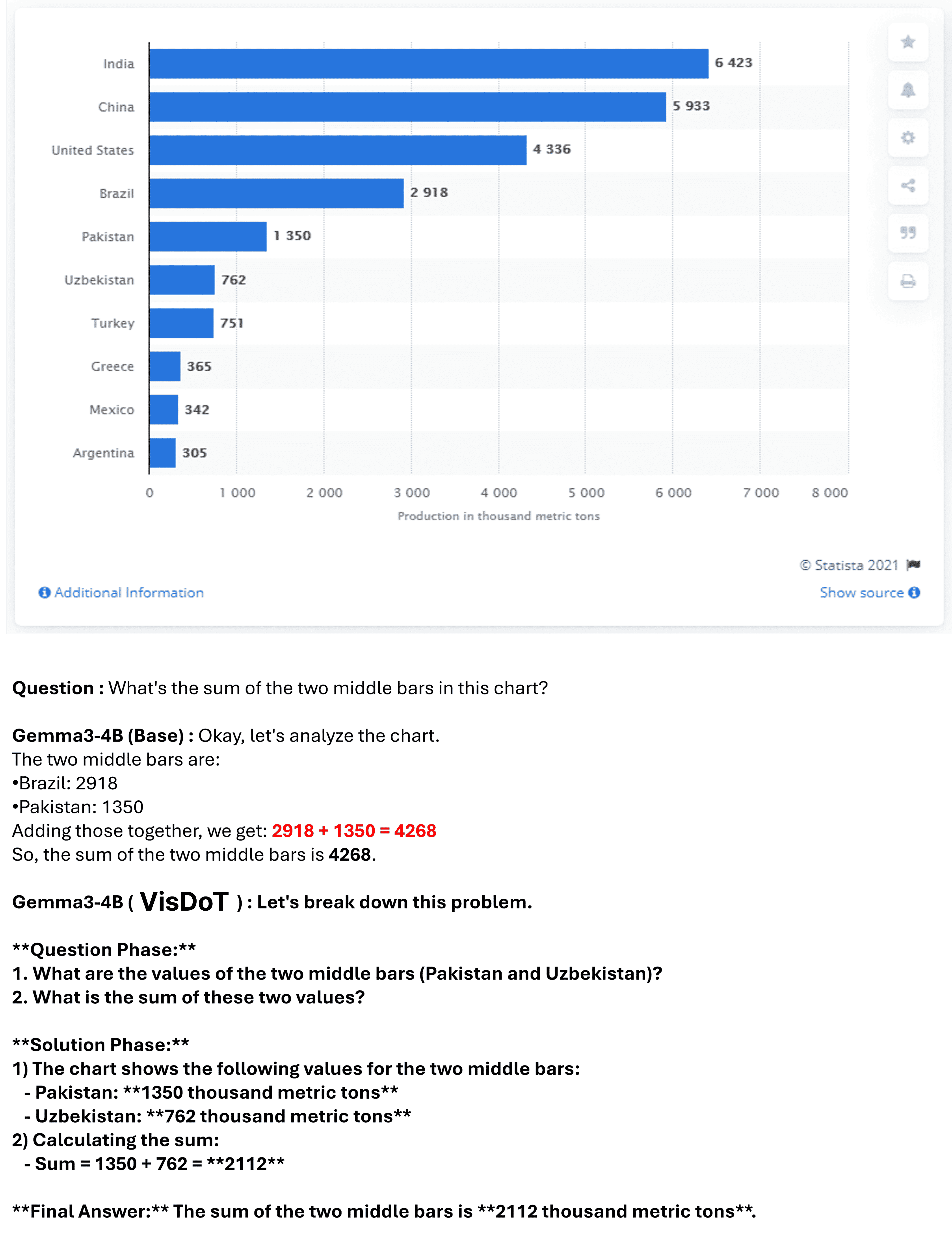}
    \caption{ChartQA Human task 2}
    \label{fig:chartqa2}
\end{figure*}

\begin{figure*}[!t]
    \centering
    \includegraphics[width=\textwidth]{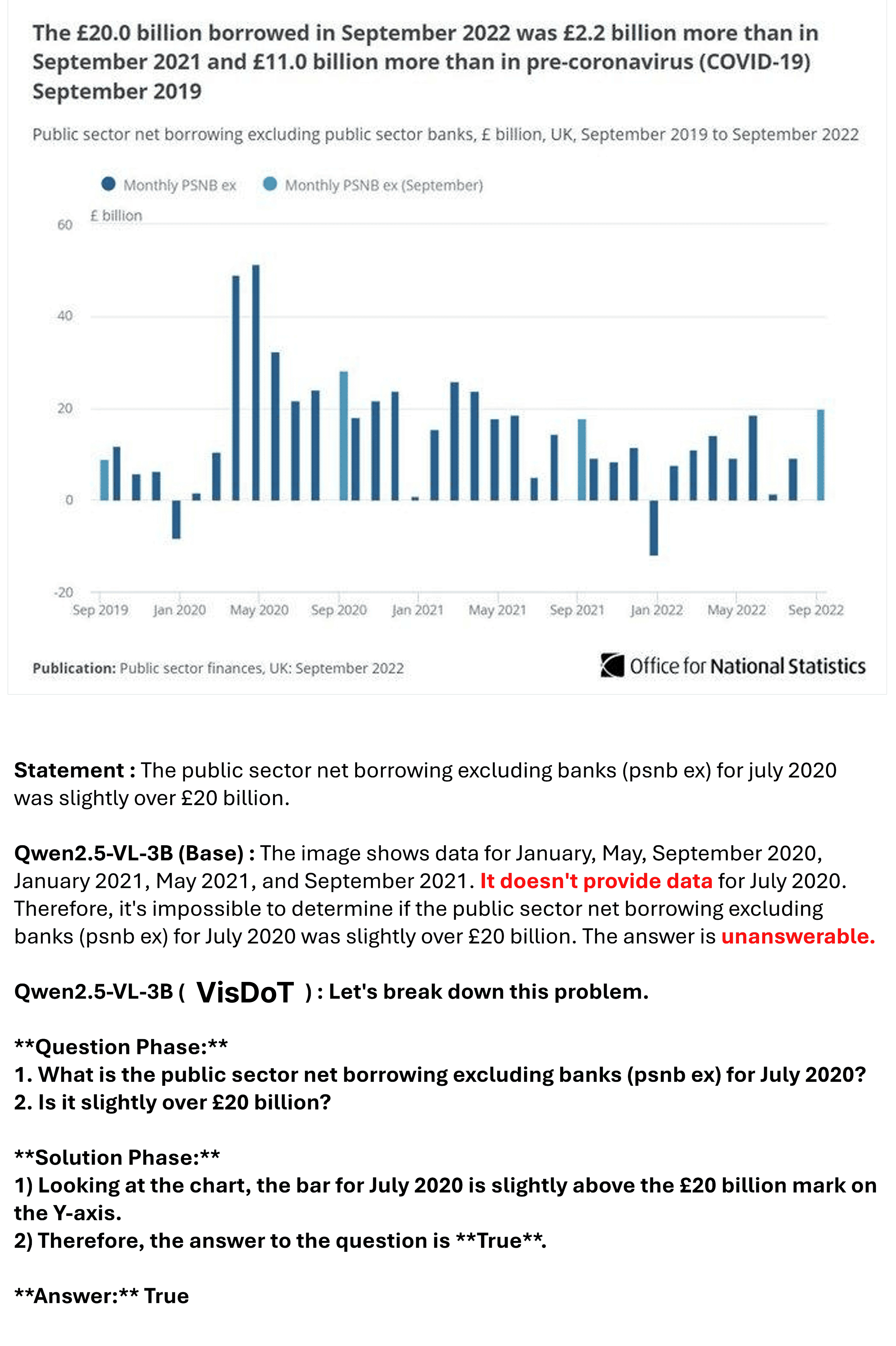}
    \caption{ChartQAPro Fact Checking task}
    \label{fig:chartqapro_check}
\end{figure*}

\begin{figure*}[!t]
    \centering
    \includegraphics[width=\textwidth]{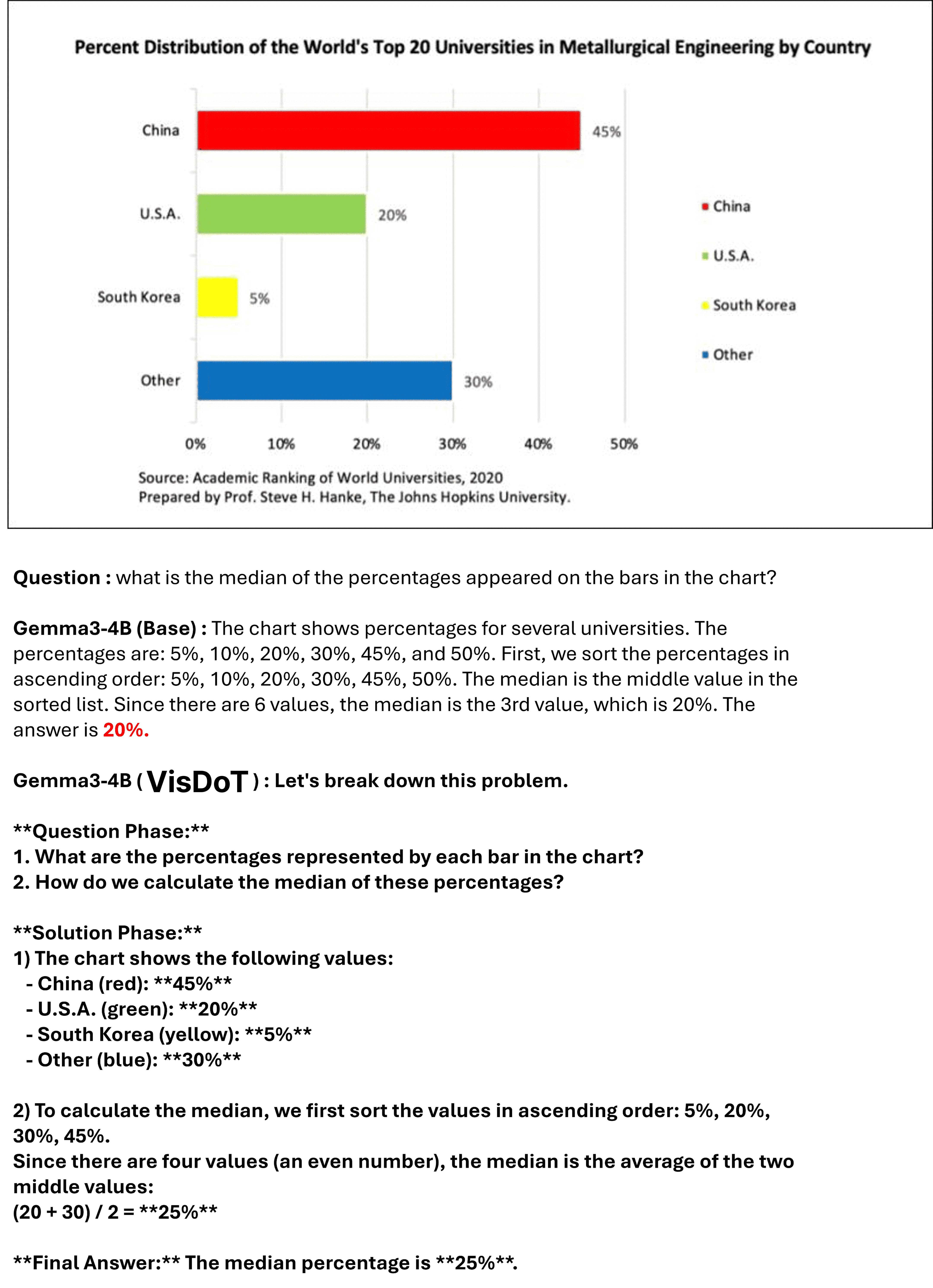}
    \caption{ChartQAPro Factoid task}
    \label{fig:chartqapro_factoid}
\end{figure*}

\begin{figure*}[!t]
    \centering
    \includegraphics[width=\textwidth]{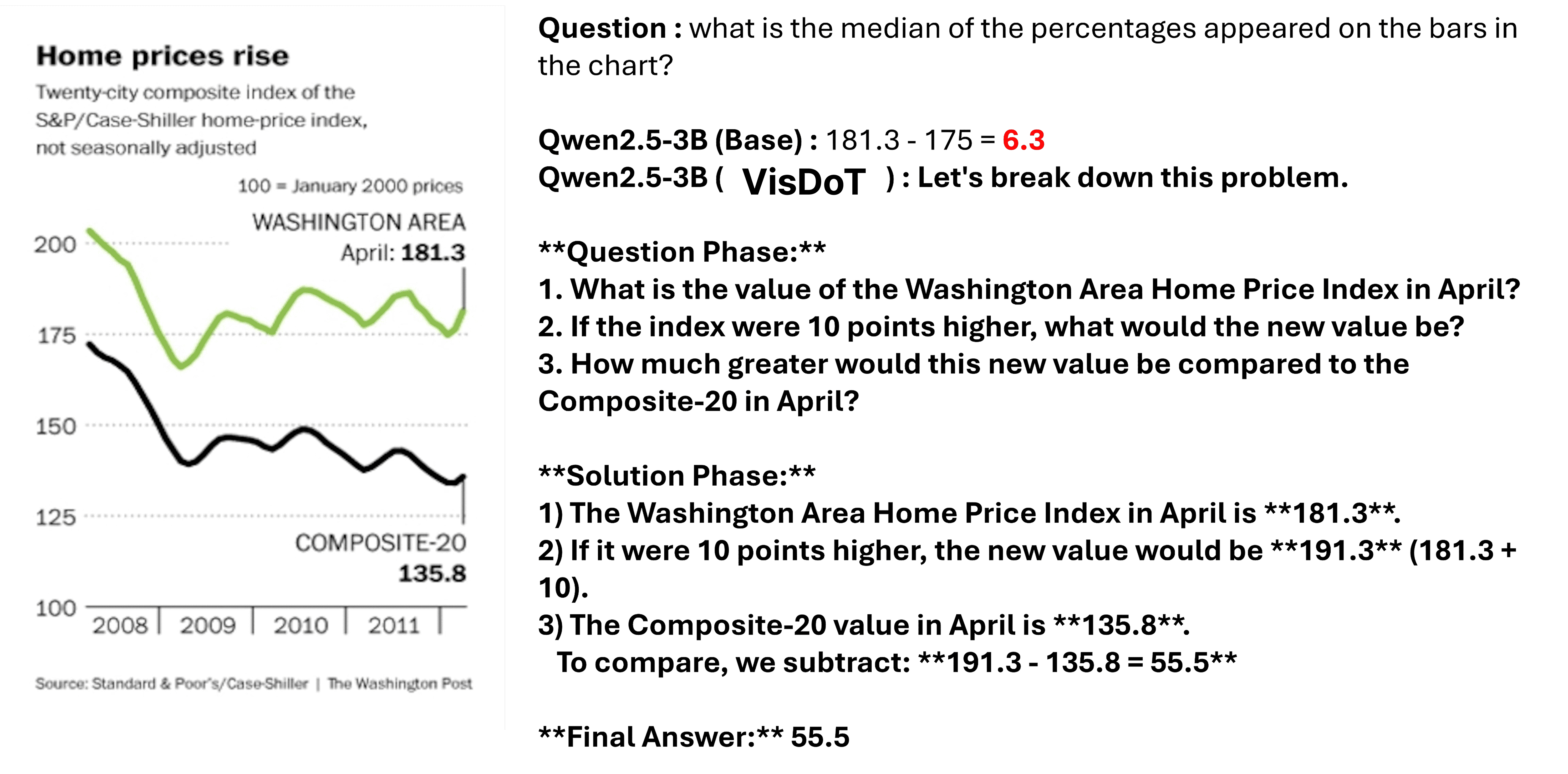}
    \caption{ChartQAPro Hypothetical task}
    \label{fig:chartqapro_hypo}
\end{figure*}

\begin{figure*}[!t]
    \centering
    \includegraphics[width=\textwidth]{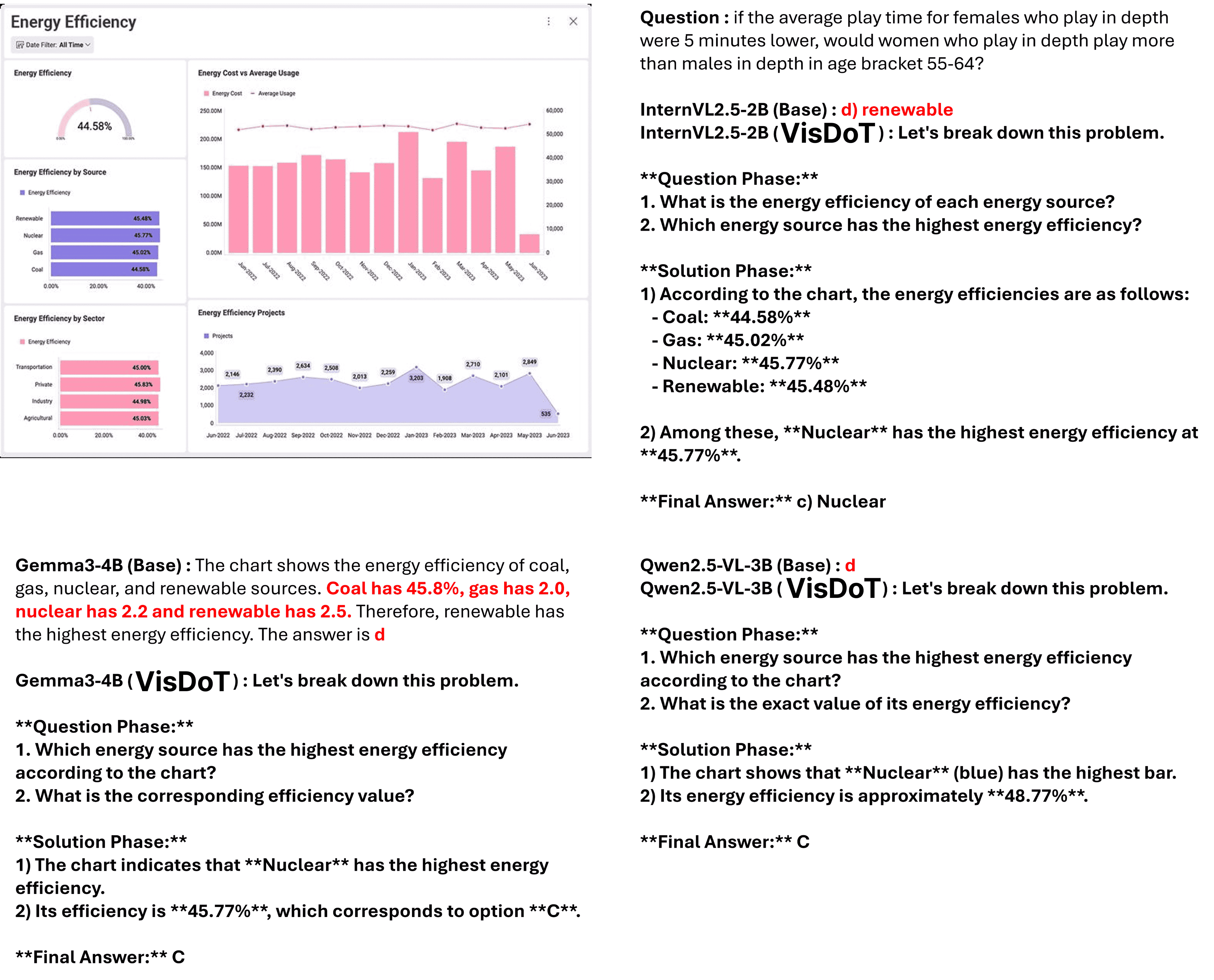}
    \caption{ChartQAPro Multi Choice task}
    \label{fig:chartqapro_multi}
\end{figure*}

\begin{figure*}[!t]
    \centering
    \includegraphics[width=\textwidth]{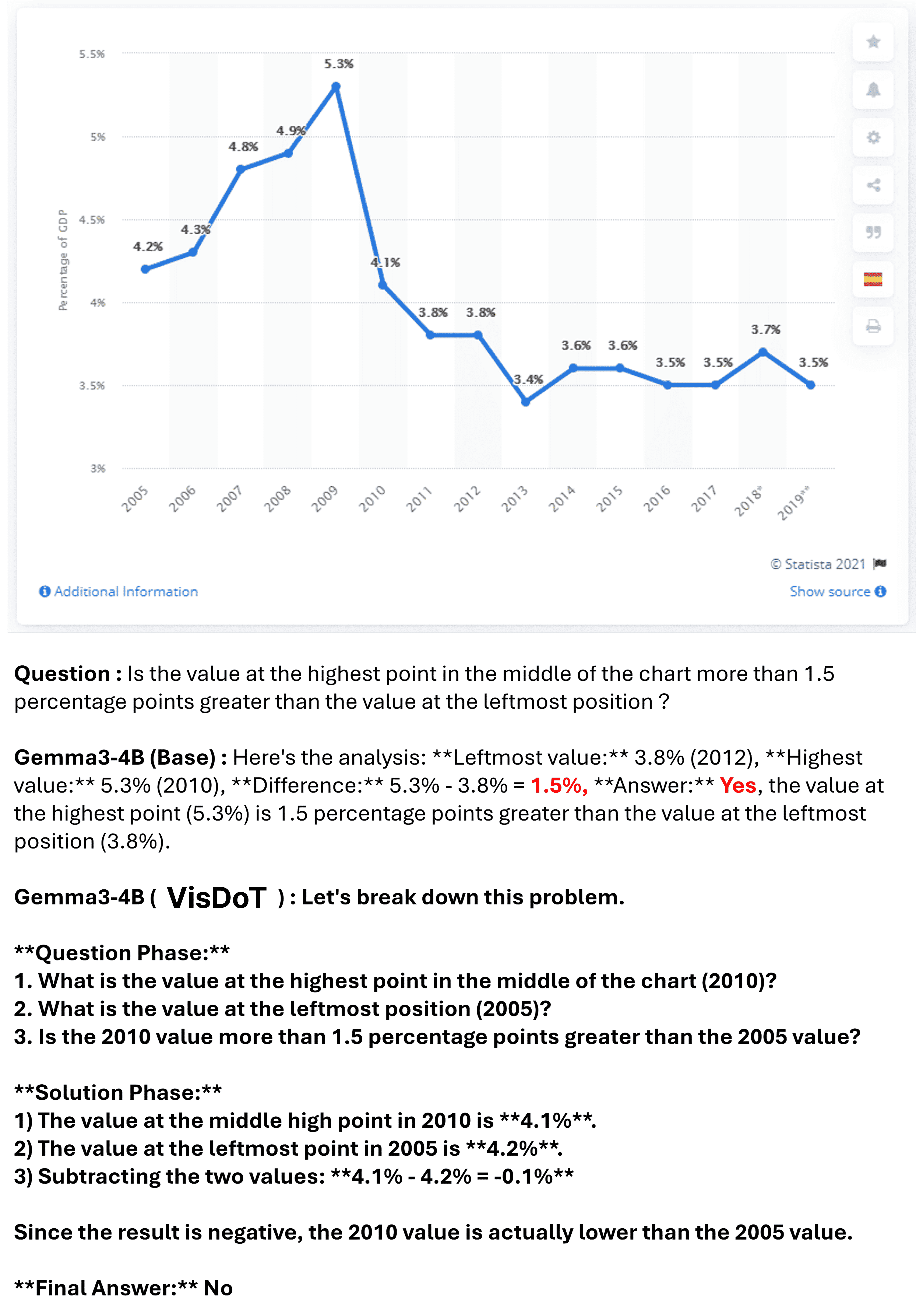}
    \caption{VisDoTQA Position task} 
    \label{fig:visdot_position}
\end{figure*}

\begin{figure*}[!t]
    \centering
    \includegraphics[width=\textwidth]{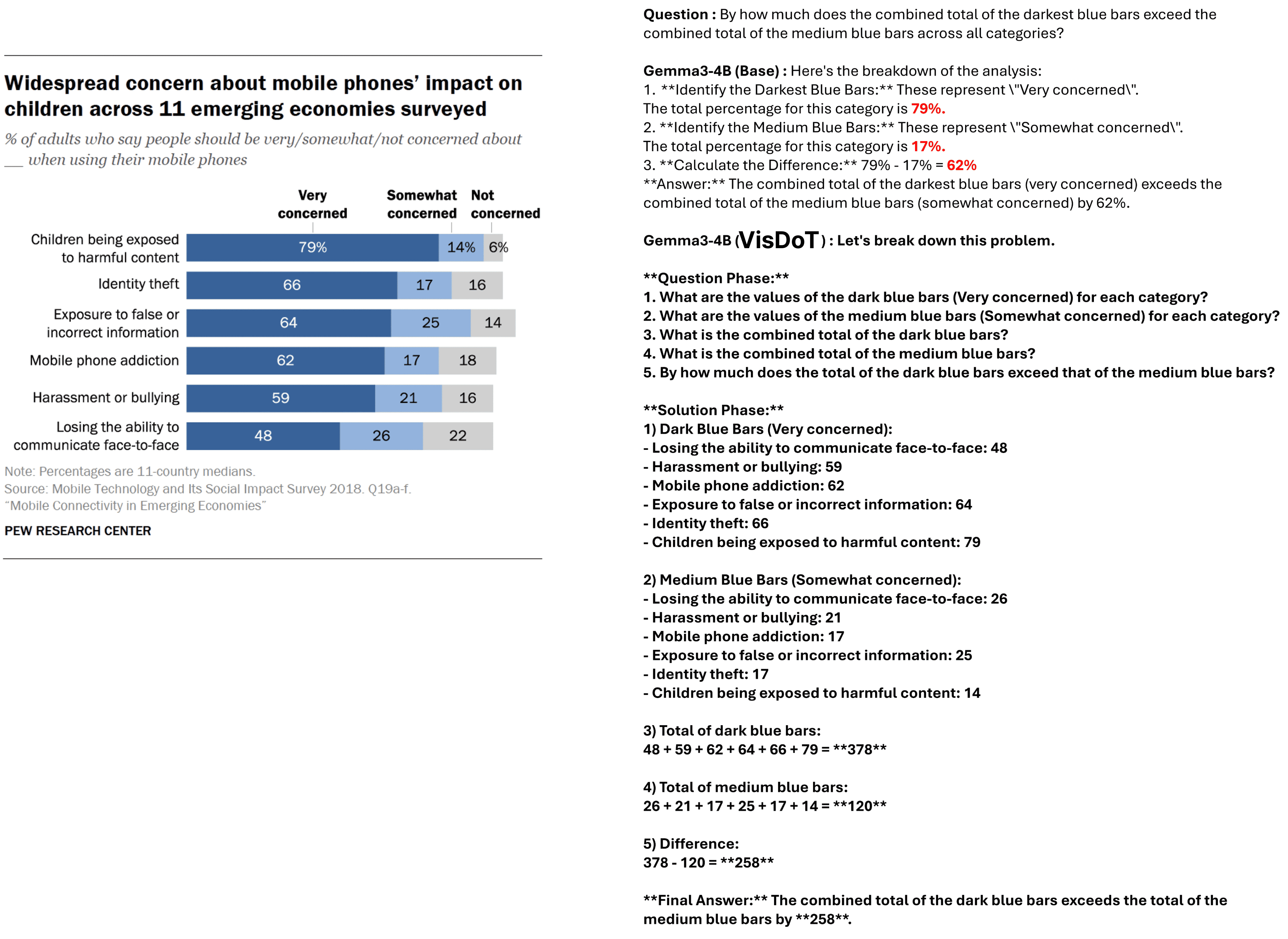}
    \caption{VisDoTQA Pattern task}
    \label{fig:visdot_pattern}
\end{figure*}

\begin{figure*}[!t]
    \centering
    \includegraphics[width=\textwidth,height=0.82\textheight,keepaspectratio]{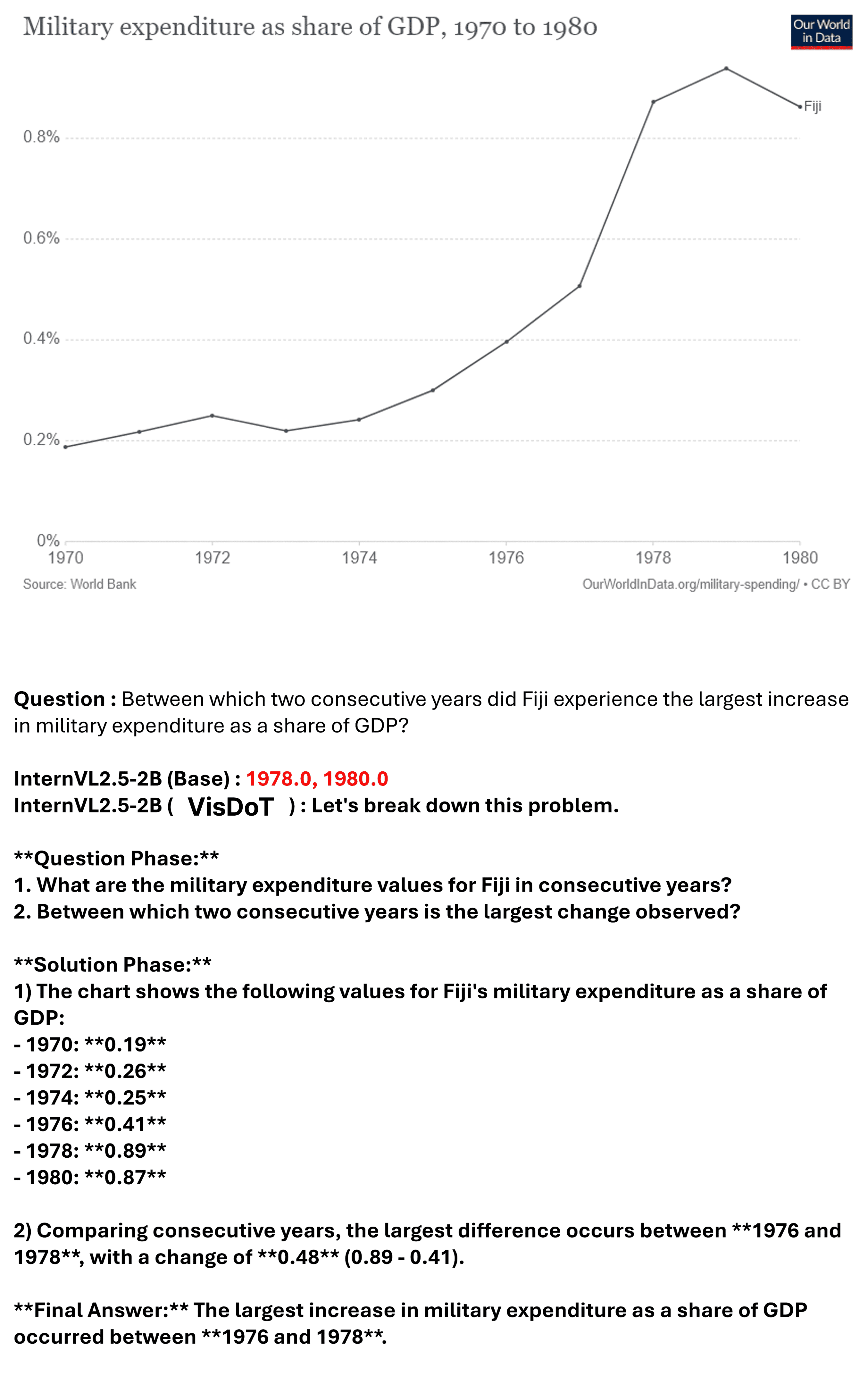}
    \caption{VisDoTQA Length task}
    \label{fig:visdot_length}
\end{figure*}

\begin{figure*}[!t]
    \centering
    \includegraphics[width=\textwidth]{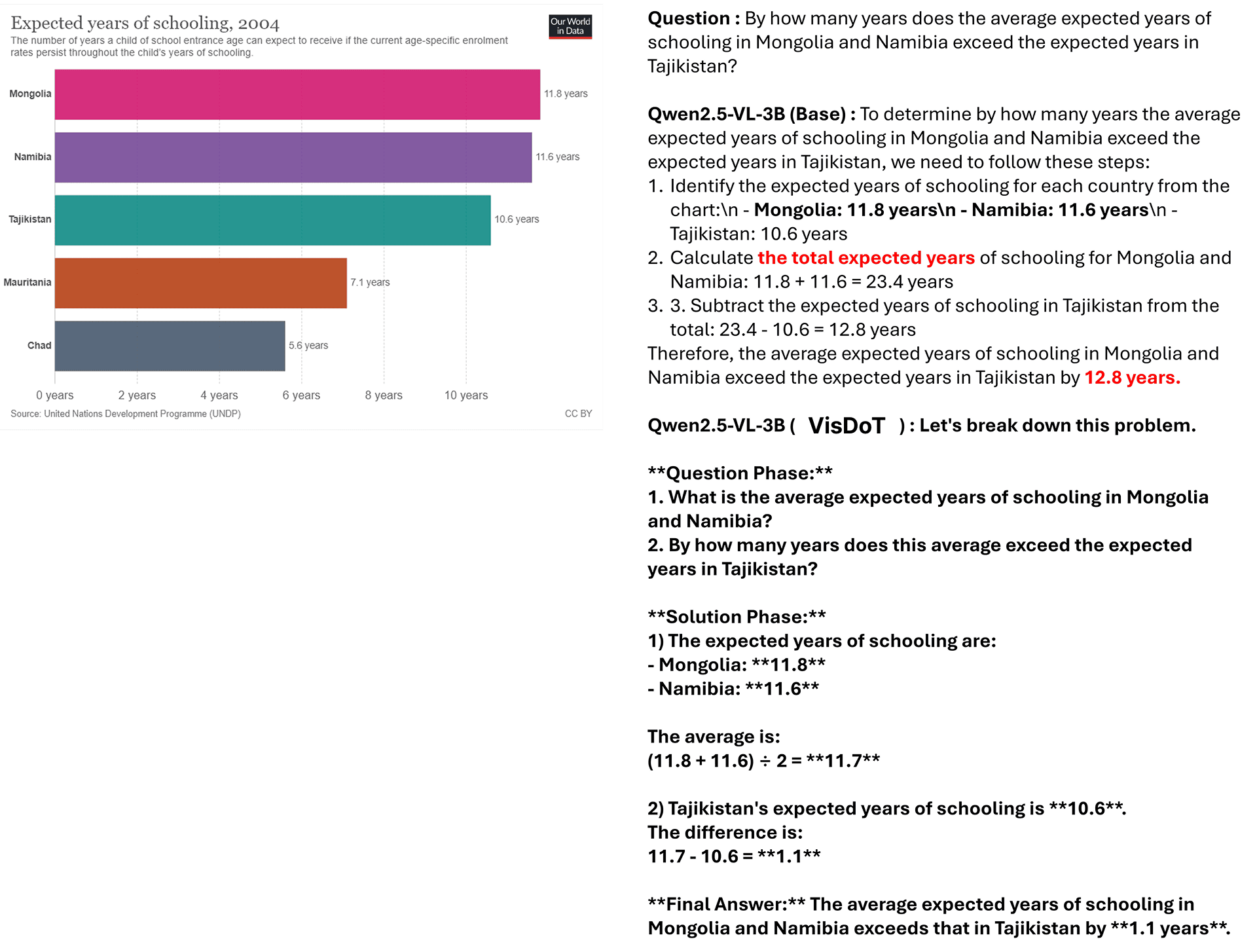}
    \caption{VisDoTQA Extract task}
    \label{fig:visdot_extract}
\end{figure*}

\subsection{Evaluation Case Studies}
\label{appendix:c.4}
In this section, we present case studies demonstrating how fine-tuned models effectively respond to chart-based questions. Representative examples are selected from the ChartQA, ChartQAPro, and VisDoTQA human split subsets. These examples highlight the enhanced reasoning capabilities achieved through fine-tuning. 

Our analysis focuses on the outputs of fine-tuned models, as baseline models often fail to produce verifiable or coherent answers in complex chart reasoning tasks. Each example illustrates the model’s ability to accurately recognize visual cues and perform multi-step reasoning to arrive at the correct answer. 

Figure~\ref{fig:chartqa1},~\ref{fig:chartqa2}: Results on the ChartQA Human task

Figure~\ref{fig:chartqapro_check},~\ref{fig:chartqapro_factoid},~\ref{fig:chartqapro_hypo},~\ref{fig:chartqapro_multi}: Results on the ChartQAPro task

Figure~\ref{fig:visdot_position},~\ref{fig:visdot_pattern},~\ref{fig:visdot_length},~\ref{fig:visdot_extract}: Results on the VisDoTQA task

To better understand how models handle complex chart-based questions, we analyzed responses on VisDoTQA using a unified prompt template. This comparison reveals clear behavioral differences between baseline and fine-tuned models. 

The InternVL2.5 baseline often generated short, direct answers and failed to follow the multi-step reasoning structure provided in the prompt. Even when structured guidance was available, the model tended to return a single numerical value without explaining the reasoning process. This indicates difficulties in comprehending complex questions and producing accurate answers. 

The Qwen2.5-VL-3B baseline showed a stronger tendency toward step-by-step reasoning, aligning with its general instruction-following capabilities. However, it frequently misinterpreted visual elements (e.g., bar heights or values) and exhibited weak numerical reasoning, leading to unit confusion or incorrect calculations. These errors suggest that while the model attempts structured inference, it lacks the visual grounding and numerical precision required for chart-based reasoning. 

In contrast, all models fine-tuned on the VisDoTQA dataset consistently followed the reasoning structure outlined in the prompt. Despite using the same system prompt as the baselines, the fine-tuned models successfully decomposed complex questions, accurately interpreted visual content, and generated correct answers through sequential reasoning. This demonstrates that VisDoT-based fine-tuning significantly improves a model’s ability to align visual perception with logical inference, even without additional prompt engineering.

\section{Detailed Prompt}
\label{appendix:d}

\subsection{Perception-following Question Generation}
\label{appendix:d.1}
Below are the prompt templates used for generating Perception-following Questions. Each template is designed to elicit questions corresponding to one of the four perceptual tasks: Position, Length, Pattern, and Extract. The full prompt is presented in Table~\ref{tab:perception_prompt_position}, ~\ref{tab:perception_prompt_length},~\ref{tab:perception_prompt_pattern} and~\ref{tab:perception_prompt_extract}.

\begin{table*}[t]
\small
\centering
\begin{tabularx}{\textwidth}{@{}X@{}}
\toprule
\textbf{System Prompt: Question Generation - Position Prompt} \\
\midrule
You are an AI assistant that generates \textbf{location-based analytical questions} from structured visual data, such as charts and tables.  
Your task is to analyze the image and generate \textbf{questions that extract values based on their spatial position}. \\
\\
\textbf{How to Generate Questions} \\
- Use \textbf{the relative position of elements} instead of direct category names as the primary reference. \\
- However, \textbf{also include the actual name/label in parentheses} after each positional reference for clarity. \\
- Example: \\
\quad - \textbf{Rows (Vertical Order)}: "third region from the top (South America)", "second from the bottom (Europe)" \\
\quad - \textbf{Columns (Horizontal Order)}: "leftmost column (2019)", "second from the right (2021)" \\
\\
- Ensure a mix of different \textbf{positional references} (top, bottom, upper, lower, first, last, left, right). \\
- Avoid simple retrieval questions—each question should require some form of reasoning (comparison, sum, difference, etc.). \\
\\
\textbf{Task} \\
Generate \textbf{2 complex analytical questions} based on spatial positioning. \\
Use both \textbf{relative positions} and \textbf{actual names/labels} in parentheses. \\
\bottomrule
\end{tabularx}
\caption{Perception-following Question Generation - Position Prompt}
\label{tab:perception_prompt_position}
\end{table*}


\begin{table*}[t]
\small
\centering
\begin{tabularx}{\textwidth}{@{}X@{}}
\toprule
\textbf{System Prompt: Perception-following Question Generation - Length Prompt} \\
\midrule
System Prompt \\
You are an expert in analyzing data visualizations and generating insightful questions. \\
Based on a given chart image, generate appropriate questions. \\
\\
\textbf{Required Operations} \\
- \textbf{extract}: Ask for the value of a specific category. \\
- \textbf{ranking}: Identify the item, year, or group with the highest, 2nd, 3rd,... lowest value. \\
- \textbf{comparison}: Compare two or more values or categories. \\
- \textbf{counting}: Count how many items meet a specific condition. \\
- \textbf{addition \& subtraction}: Summing or finding differences between multiple values. \\
- \textbf{multiplication \& division}: Calculating proportional changes or percentage comparisons. \\
- \textbf{ratio}: Determining how many times one value is larger or smaller than another. \\
- \textbf{rate of change}: Identify intervals or categories based on how much their values have changed over time. \\
\\
\textbf{Output Format} \\
 "type": "extract", "question": "What was the export share in 2007?", \\
"type": "ranking", "question": "Which month had the lowest hiring rate?", \\
"type": "comparison", "question": "Compare the hiring rates of May 2020 and February 2021.", \\
"type": "comparison", "question": "Which country data is consistently above 1.5 kg?", \\
"type": "rate of change", "question": "Which two consecutive years showed the largest change?", \\
"type": "subtraction", "question": "How did the export share change from 2006 to 2009?", \\
"type": "comparison + counting", "question": "How many years had a higher export share than 2010?", \\
"type": "subtraction + ranking", "question": "What is the difference between the highest and lowest export share years?"  \\
\bottomrule
\end{tabularx}
\caption{Perception-following Question Generation - Length Prompt}
\label{tab:perception_prompt_length}
\end{table*}

\begin{table*}[t]
\small
\centering
\begin{tabularx}{\textwidth}{@{}X@{}}
\toprule
\textbf{System Prompt: Perception-following Question Generation - Pattern Prompt} \\
\midrule
You are an AI assistant that generates analytical questions based on recurring visual patterns in charts.  
Your task is to create \textbf{complex, meaningful questions} that require pattern recognition and reasoning over visual groupings such as colors, legends, markers, or labeled categories, rather than simple label-based lookup. \\
\\
\textbf{Required Operations (use each at least twice across 20 questions):} \\
- \textbf{Pattern Counting}: Identifying how many elements share a common visual pattern (e.g., “How many wedges are in shades of blue?”). \\
- \textbf{Color/Legend Matching}: Comparing values between categories based on shared legend color or visual symbol. \\
- \textbf{Pattern-based Addition \& Subtraction}: Summing or comparing values across elements sharing the same visual style (e.g., line types, bar textures). \\
- \textbf{Ratio}: Determining the proportional relationship between visually grouped categories. \\
- \textbf{Comparison}: Identifying which visually marked group has greater or lesser values. \\
- \textbf{Pattern Ranking}: Ordering patterns by magnitude or frequency of values (e.g., which color category has the 2nd highest value). \\
\\
\textbf{Important Guidelines} \\
- Clearly include \textbf{at least two reasoning operations} per question. \\
- \textbf{All listed operations} must appear at least \textbf{twice across the 20 questions}. \\
- Avoid surface-level or pattern-mention-only questions—require true reasoning over the visual encodings. \\
- Focus on pattern-informed reasoning such as matching legends, interpreting repeated visual cues, or comparing across symbol groups. \\
\\
\textbf{Example Questions} \\
1. What is the total percentage of the three categories shaded in red? (Pattern Counting, Addition) \\
2. Which legend color represents the category with the largest difference from the average? (Legend Matching, Comparison) \\
3. How much higher is the value for the striped bar than the dotted bar? (Pattern Comparison, Subtraction) \\
4. Which marker shape corresponds to the second-highest value? (Visual Pattern Ranking) \\
5. Among all categories with similar shades of green, which has the lowest count? (Color Grouping, Ranking) \\
6. How many segments share both the same color and bar direction? (Pattern Matching, Counting) \\
\bottomrule
\end{tabularx}
\caption{Perception-following Question Generation - Pattern Prompt}
\label{tab:perception_prompt_pattern}
\end{table*}


\begin{table*}[t]
\small
\centering
\begin{tabularx}{\textwidth}{@{}X@{}}
\toprule
\textbf{System Prompt: Perception-following Question Generation - Extract Prompt} \\
\midrule
You are an AI assistant that generates analytical questions based on numerical data in charts. Your task is to create \textbf{complex, meaningful questions} that require reasoning and analysis rather than simple data lookup. \\
\\
\textbf{Required Operations (use each at least twice across 20 questions):} \\
- \textbf{Counting}: Identifying how many items meet a specific numerical condition (e.g., "How many regions exceed 20\%?"). \\
- \textbf{Addition \& Subtraction}: Summing or finding differences between multiple values. \\
- \textbf{Multiplication \& Division}: Calculating proportional changes or percentage comparisons. \\
- \textbf{Average \& Median}: Comparing means and middle values across multiple categories or regions. \\
- \textbf{Ratio}: Determining how many times one value is larger or smaller than another. \\
- \textbf{Ranking}: Clearly identifying positions (highest, lowest, top 3, bottom 2, etc.). \\
- \textbf{Comparison}: Identifying higher/lower values or categories without performing explicit mathematical operations. \\
\\
\textbf{Important Guidelines} \\
- Clearly state \textbf{at least two operations} per question. \\
- \textbf{All listed operations} must be used at least \textbf{twice across the 20 questions}. \\
- Avoid overly simple or single-operation arithmetic questions. \\
- Ensure each question encourages \textbf{analytical thinking and deeper interpretation} of chart data. \\
\\
\textbf{Example Questions} \\
1. What is the combined percentage of the two highest-ranked categories? (\textit{Addition, Ranking}) \\
2. By what percentage is the average of Category A higher than the median of Category B? (\textit{Average, Median, Subtraction}) \\
3. How much higher is the 2nd highest value than the 4th highest value in the Sales column? (\textit{Ratio, Ranking}) \\
4. Which three categories have the lowest values in the Good rating column? (\textit{Ranking}) \\
5. Which country has the largest gap between negative and positive ratings? (\textit{Subtraction, Ranking}) \\
6. How many regions have a value above the overall average in the Bachelor’s degree category? (\textit{Counting, Average, Comparison}) \\
\bottomrule
\end{tabularx}
\caption{Perception-following Question Generation - Extract Prompt}
\label{tab:perception_prompt_extract}
\end{table*}

\subsection{Decomposition-of-Thought Prompt}
\label{appendix:d.2}
We present the full DoT prompting template referenced in Section~\ref{Sec:3.2}. The prompt first instructs the model to decompose a given question into sub-questions, explicitly prioritizing the generation of perception-oriented sub-questions before cognitive ones. Subsequently, the model is guided to sequentially generate intermediate answers for each sub-question, culminating in a final answer that integrates the accumulated reasoning steps. The full prompt is presented in Table~\ref{tab:dot_prompt}.

\begin{table*}[t]
\small
\centering
\begin{tabularx}{\textwidth}{@{}X@{}}
\toprule
\textbf{System Prompt: DoT Prompt} \\
\midrule
You are an AI assistant that analyzes chart images and constructs the reasoning process to derive answers when given question–answer pairs. \\
You must follow a structured approach by decomposing the question into sub-questions and solving them systematically. \\
\\
\textbf{Response Format:} \\
--- \\
Let’s break down this problem. \\
\textbf{Question Phase:} \\
1. (Ask about element 1) \\
2. (Ask about element 2) \\
3. ... \\
\textbf{Solution Phase:} \\
--- \\
1) (Answer sub-question 1 with reasoning and answer) \\
2) (Answer sub-question 2 with reasoning and answer) \\
3) ... \\
\\
\textbf{Guidelines:} \\
--- \\
\textbf{- Question Phase:} \\
\quad • Identify the necessary elements from the question to solve the problem. \\
\quad • Generate sub-questions \textbf{strictly following the format}: \\
\quad \quad 1. (Ask about element 1) \\
\quad \quad 2. (Ask about element 2) \\
\quad \quad 3. ... \\
\quad • The number of sub-questions must be \textbf{exactly reflected in the Solution Phase}. \\
\\
\textbf{- Solution Phase:} \\
\quad • Ensure that every sub-question has a corresponding answer in the exact order. \\
\quad • Provide answers \textbf{strictly following the format}: \\
\quad \quad 1) (Answer sub-question 1 with reasoning and answer) \\
\quad \quad 2) (Answer sub-question 2 with reasoning and answer) \\
\quad \quad 3) ... \\
\quad • Responses must be \textbf{clear, precise, and follow the structured format without skipping or merging answers}. \\
\quad • \textbf{Do not include the word "table" in any response.} \\
\bottomrule
\end{tabularx}
\caption{DoT Prompt}
\label{tab:dot_prompt}
\end{table*}

\subsection{Model Inference Prompt}
\label{appendix:d.3}

We present the inference prompts used across various benchmarks and experimental settings. All prompts were provided to the model as system-level instructions. The full prompt is presented in Table~\ref{tab:model_inference_prompt}.

\begin{table*}[t]
\small
\centering
\begin{tabularx}{\textwidth}{@{}l|X@{}}
\toprule
\textbf{Category} & \textbf{Prompt Template} \\
\midrule
\textbf{ChartQAPRO (fact checking)} & 
You are a meticulous chart-analysis assistant.  
Based on the chart image, decide whether the following statement is true or false.  
Your final answer must be either ‘True’ or ‘False’. \\
\midrule
\textbf{ChartQAPRO (multi choice)} & 
You are a meticulous chart-analysis assistant.  
Read the chart image and select the most appropriate answer  
from the multiple-choice options in the question,  
returning only the final answer as a single letter: A, B, C, or D. \\
\midrule
\textbf{ChartQAPRO (hypothetical, factoid)} & 
You are a meticulous chart-analysis assistant.  
Analyze the chart image and answer the following question. \\
\midrule
\textbf{ChartQAPRO (conversational)} & 
You are a meticulous chart-analysis assistant.  
Your goal is to read the multi-turn conversation carefully and provide the answer to the final question.  
Conversation:  
\{conversation\_text\}  
Question: \{final\_question\} \\
\midrule
\textbf{ChartQA, VisDoT} & 
You are a meticulous chart-analysis assistant.  
Analyze the chart image and answer the following question. \\
\bottomrule
\end{tabularx}
\caption{Model Inference Prompt}
\label{tab:model_inference_prompt}
\end{table*}

\section{Representative Examples of the VisDoTQA Dataset}
\label{appendix:e}

To qualitatively illustrate the diversity and reasoning patterns captured by our VisDoTQA dataset, we present four representative examples, each corresponding to a distinct type of reasoning: \textbf{Position-based}, \textbf{Length-based}, \textbf{Pattern-based}, and \textbf{Extractive}.

\paragraph{Data Source.}
All chart images used in this appendix are sourced from the publicly available \textbf{ChartQA} dataset~\cite{masry2022chartqa}. The visual content remains unchanged, while the questions and answers are newly constructed based on our proposed DoT-guided reasoning framework.

\paragraph{(1) Position-based Reasoning}
\begin{itemize}
    \item \textbf{Question:} Rank the regions based on their dark purple bar revenue, and determine how much higher the top-ranked region's revenue is compared to the bottom-ranked region.
    \item \textbf{Answer:} The dark purple (2020) revenues are: The Americas (565,023), Europe (326,613), Japan (301,187), Other (115,694). The difference between the top and bottom regions is \(565{,}023 - 115{,}694 = \textbf{449{,}329}\).
\end{itemize}

\begin{figure*}[h]
    \centerline{\includegraphics[width=\textwidth]{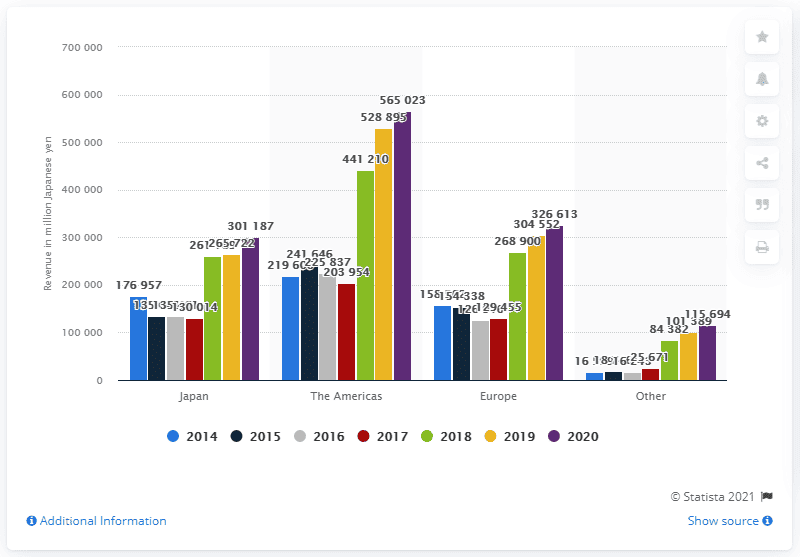}}
    \caption{Example of position-based reasoning.}
    \label{fig:Position}
\end{figure*}

Figure~\ref{fig:Position} illustrates this position-based reasoning example.

\paragraph{(2) Length-based Reasoning}
\begin{itemize}
    \item \textbf{Question:} Find the median time for trajectory in black between 2010 and 2019, and compare it to the median time for blue trend.
    \item \textbf{Answer:} For the black line (Women), the median is 233.5 minutes; for the blue line (Men), it is 203 minutes. Thus, the black trend is longer by \textbf{30.5 minutes}.
\end{itemize}

\begin{figure*}[h]
    \centerline{\includegraphics[width=\textwidth]{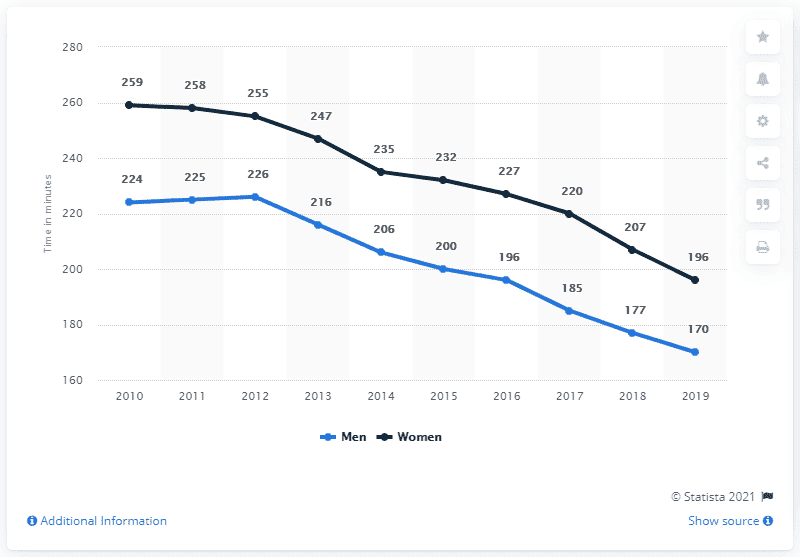}}
    \caption{Example of length-based reasoning.}
    \label{fig:length}
\end{figure*}

Figure~\ref{fig:length} shows the corresponding length-based reasoning instance.

\begin{figure*}[h]    \centerline{\includegraphics[width=\textwidth]{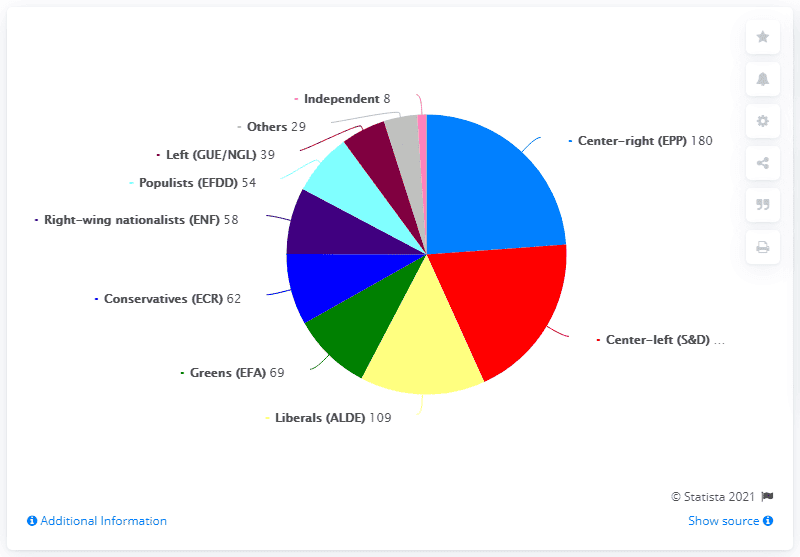}}
    \caption{Example of pattern-based reasoning.}
    \label{fig:pattern}
\end{figure*}

\paragraph{(3) Pattern-based Reasoning}
\begin{itemize}
    \item \textbf{Question:} Identify the group that holds about one-fourth of the total seats occupied by the five largest groups combined.
    \item \textbf{Answer:} The total seats of top-5 groups = 566. One-fourth is approximately 141.5. The Center-left (S\&D) group holds 146 seats, which is closest. \textbf{Answer: Center-left (S\&D)}.
\end{itemize}
Figure~\ref{fig:pattern} demonstrates the pattern-based reasoning task.

\begin{figure*}[h]    \centerline{\includegraphics[width=\textwidth]{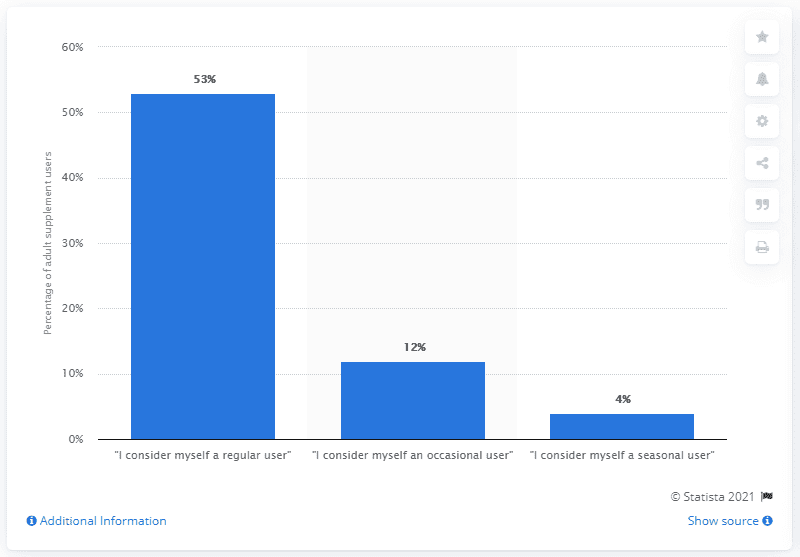}}
    \caption{Example of extractive reasoning.}
    \label{fig:extractive}
\end{figure*}

\paragraph{(4) Extractive Reasoning}
\begin{itemize}
    \item \textbf{Question:} Calculate the difference between the highest and lowest percentage categories.
    \item \textbf{Answer:} Highest = 53\% (Regular user), Lowest = 4\% (Seasonal user). Difference = \(53 - 4 = \textbf{49\%}\).
\end{itemize}
Figure~\ref{fig:extractive} presents the extractive reasoning case.

\end{document}